\documentclass{article}

\usepackage[preprint]{neurips_2025}

\usepackage[utf8]{inputenc} 
\usepackage[T1]{fontenc}    
\usepackage{hyperref}       
\usepackage{url}            
\usepackage{booktabs}       
\usepackage{amsfonts}       
\usepackage{nicefrac}       
\usepackage{microtype}      
\usepackage{xcolor}         
\usepackage{amsmath}
\usepackage{graphicx}
\usepackage{subcaption}
\usepackage[ruled,vlined,noend]{algorithm2e}

\SetCommentSty{mycommfont}
\DeclareMathOperator*{\argmin}{arg\,min}
\usepackage{enumitem}
\usepackage{dialogue}
\usepackage{listings}
\usepackage{tabularx}
\usepackage[most]{tcolorbox}
\definecolor{dialoguebg}{RGB}{240,248,255} 
\newtcolorbox{coloreddialogue}{
	colback=dialoguebg,
	colframe=white,
	boxrule=0.1pt,
	arc=4pt,
	left=1pt, right=1pt, top=2pt, bottom=2pt
}

\newcommand{\BC} {B\'{e}zier curve}
\newcommand{\DefInt}[4]{\ensuremath{\int\limits_{#1}^{#2} {#3} \; \mathrm{d} {#4} }}
\newcommand{\CP}{\ensuremath{\mathbf p}}
\newcommand{\CPi}[1]{\ensuremath{\CP_{#1}}}
\newcommand{\T}{\ensuremath{\tau_\theta(u)}}
\newcommand{\Td}{\ensuremath{\dot \tau_\theta(u)}}
\newcommand{\Tdd}{\ensuremath{\ddot \tau_\theta(u)}}
\newcommand{\C}[2]{\ensuremath{C_{#1,#2}}}

\newcommand{\pVar}[2]{\colorbox{#1}{#2}}
\newcommand{\pgvar}[1]{\pVar{green}{\{#1\}}}
\newcommand{\pfVar}[2]{\colorbox{#1}{\framebox{#2}}}

\usepackage{geometry}
\usepackage{graphicx}

\usepackage{amsmath}
\usepackage{listings}
\usepackage{xcolor}
\usepackage{enumitem}

\definecolor{codegreen}{rgb}{0,0.6,0}
\definecolor{codegray}{rgb}{0.5,0.5,0.5}

\definecolor{backcolour}{RGB}{245,248,250}
\definecolor{emph}{RGB}{166,88,53}
\definecolor{nightblue}{RGB}{9,49,105}
\definecolor{keywords}{RGB}{207,33,46}
\definecolor{lightpurple}{RGB}{130,81,223}

\lstdefinestyle{mystyle}{
    backgroundcolor=\color{backcolour},   
    commentstyle=\color{codegreen},
    keywordstyle=\color{keywords},
    stringstyle=\color{nightblue},
    basicstyle=\fontsize{7}{8}\ttfamily,
    breakatwhitespace=true,         
    breaklines=true,                 
    captionpos=b,                    
    keepspaces=true,                 
    numberstyle=\tiny\color{codegray},
    numbersep=2pt,                  
    showspaces=false,                
    showstringspaces=false,
    showtabs=false,                  
    tabsize=2,
    emph={dspy},
    emphstyle={\color{lightpurple}},
    linewidth=1\columnwidth,
    frame=tb,    
    xrightmargin=0pt,
    xleftmargin=0.23cm,
    numbers=left,
    aboveskip=0.2cm,
    belowskip=0.1cm,
}

\SetAlFnt{\small}
\SetAlCapFnt{\small}
\SetAlCapNameFnt{\small}
\SetAlCapHSkip{0pt}

\title{xInv: Explainable Optimization of Inverse Problems}

\author{%
  Sean Memery \\
  School of Informatics\\
  University of Edinburgh\\
  Scotland \\
  \texttt{s.memery@ed.ac.uk} \\
  \And
  Kevin Denamganai \\
  School of Informatics\\
  University of Edinburgh\\
  Scotland \\
  \texttt{Kevin.Denamganai@ed.ac.uk} \\
  \And
  Anna Kapron-King \\
  School of Informatics\\
  University of Edinburgh\\
  Scotland \\
  \texttt{A.Kapron-King@ed.ac.uk} \\
  \And
  Kartic Subr \\
  School of Informatics\\
  University of Edinburgh\\
  Scotland \\
  \texttt{K.Subr@ed.ac.uk} \\
}

\begin{document}

\maketitle

\begin{abstract}
Inverse problems are central to a wide range of fields, including healthcare, climate science, and agriculture. They involve the estimation of inputs, typically via iterative optimization, to some known forward model so that it produces a desired outcome. Despite considerable development in the explainability and interpretability of forward models, the iterative optimization of inverse problems remains largely cryptic to domain experts.  We propose a  methodology to produce explanations, from traces produced by an optimizer, that are interpretable by humans at the abstraction of the domain. The central idea in our approach is to instrument a differentiable simulator so that it emits natural language events during its forward and backward passes. In a post-process, we use a Language Model to create an explanation from the list of events. We demonstrate the effectiveness of our approach with an illustrative optimization problem and an example involving the training of a neural network.  

\end{abstract}

\section{Introduction}

The problem of estimating inputs (or parameters) that elicit a specified response from a simulation model has enjoyed much attention from mathematicians and computer scientists. These \emph{inverse problems} are typically solved via iterative updates of hypothesized input (or parameter) values. At each iteration, the outcome of the resulting \emph{forward simulation} is used to update the values and the procedure is repeated until convergence. The overall process remains largely opaque and does not lend itself to explanation regardless of the underlying optimization procedure used for the updates. The generation of explanations involves a tedious analysis procedure involving the domain expert, an expert in optimization and a detailed trace from the iterative optimizer.   We propose a simple methodology to enable human-centric solution of inverse problems using traditional optimizers.

Explainable optimization can be achieved using a variety of methods. In some (rare) cases, such as rule-based methods~\cite{freytag1987rule} or symbolic regression~\cite{Angelis2023ArtificialII}, the optimization scheme is designed to inherently lend itself to explanation. Sometimes, even if the optimizer does not permit this, it is possible to construct surrogates using methods such as model distillation~\cite{craven1995extracting, liu2018improvinginterpretabilitydeepneural} or response surface~\cite{Box1951OnTE, myers1995response}. Failing these, post hoc explanation techniques are employed including analysis of feature importance~\cite{lundberg2017unifiedapproachinterpretingmodel}, sensitivity~\cite{saltelli2004sensitivity} and visualization~\cite{ggplot2}. Our method aligns best with post hoc methods, with the added benefits of considering a human-in-the-loop due to its interactivity.  Modern Language Models (LM) are the obvious choice for explainability, especially when their knowledge of the application domain can be leveraged. However, they are infamously unreliable as optimizers or even at interpreting simulation traces containing numerical values~\cite{cherian2024llmphycomplexphysicalreasoning}.

\begin{figure}
	\begin{center}
	\includegraphics[width=\linewidth]{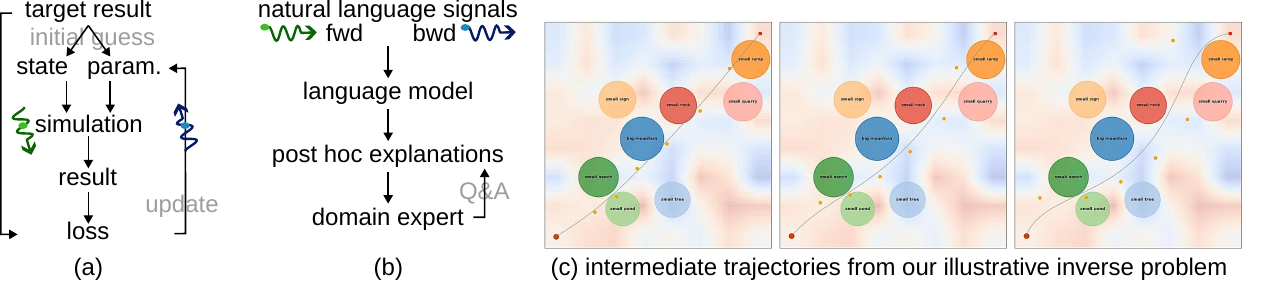}
 \caption{\label{fig:ovw} We instrument traditional inverse-solving pipelines (a) to emit two natural language signals, during the forward (simulation) and backward (differentiation) passes. (b) This enables post hoc interpretability via Language Models. We demonstrate this (c) using a  simple, but representative, optimization problem (Section~\ref{subsec:illustrative-optimization-problem}) and on optimizing (training) a small language model (Section~\ref{sec:lm_training}). }
\end{center}
\end{figure}

The key idea we introduce in this paper is to instrument the optimization process to generate a secondary trace in natural language, both during its forward and backward (differentiable) paths (see Figure~\ref{fig:ovw}). 
This involves modifying the optimizer to emit three signals: events, rewards and updates. This step will, in principle, be done by the optimization expert. Then, we define transformation functions that 
map these signals to natural language descriptions. These transformations will ideally be defined by domain experts. Finally, we implement a \emph{post hoc} scheme that exploits LMs to derive explanations from the natural language trace. We demonstrate that this simple idea provides LMs with valuable contextual information enabling interactive queries and explanation. 

We validate the efficacy of our method in two contexts. First, we present qualitative and quantitative results from an illustrative problem of optimizing a railroad, using multiple standard optimizers, in a differentiable physics simulation. Then we demonstrate the versatility of our approach by applying it to explain decisions by the optimizer while training a small LM. 

\subsection{Related Work}

Explainability with LMs~\citep{zhao2023explainabilitylargelanguagemodels, Luo2024FromUT} is a natural fit for human interepretibility, however their limitations in reasoning~\citep{kambhampati2024LLM-Modulo} and bias amplification~\citep{gallegos2024BiasAndFairness} have become apparent. LMs' explanatory deficiencies~\citep{turpin2024unfaithfulExplanationsInCoT,lanham2023measuringFaithfulness} have spurred approaches including local explanations~\citep{Ferrando2022MeasuringTM, Chuang2024LargeLM} via feature attribution, perturbation methods~\citep{Lundberg2017AUA} examining output changes after input removal, and gradient-based methods~\citep{Sundararajan2017AxiomaticAF} measuring input importance through gradient accumulation. These approaches, though insightful, lack cross-domain generalizability and suitability for online decision-making.
There is a growing body of research on explainability in optimization problems~\citep{IBS2024101297,Heaton2023, 10.1145/3701236, rachmut2025explainabledistributedconstraintoptimization, chakraborty2025explainablebayesianoptimization, 10.5555/3398761.3399089}, yet few works deliver natural language explanations or leverage differentiable simulators.
Explainability in reinforcement learning has evolved from visualizing policies and trajectories to more sophisticated approaches~\citep{Milani2023ExplainableRL,Puiutta2020ExplainableRL}. Relevant contributions include reward decomposition methods~\citep{Juozapaitis2019ExplainableRL} explaining actions as trade-offs between semantic components, and temporal policy decomposition~\citep{ruggeri2025explainablereinforcementlearningtemporal} breaking down value functions into expected future outcomes. Despite these advances, generating faithful natural language explanations for complex RL systems remains challenging.

Physics reasoning with LMs has focused on math and physics word problems~\citep{cobbe2021GSM8K,hendrycks2021MATH,lu2024mathvista,ding2023PhysQA}, with some works exploring simulator integration~\citep{Liu2022MindsEG,todorov2012mujoco,Lyu2024AdaptingWL}, and embodied environments~\citep{Wang2023VoyagerAO, fan2022minedojobuildingopenendedembodied}. Building on work in intuitive physics~\citep{baillargeon1985object,baillargeon1995physical,spelke1992origins,kim1999perception}
, systems like GRASP~\citep{jassim2024graspnovelbenchmarkevaluating} evaluate LMs' intuitive physical understanding. Studies have addressed simulation-to-reality transfer~\citep{balazadeh2024syntheticvisiontrainingvisionlanguage} and evaluated LMs in interactive environments~\citep{Mecattaf2024ALL, voudouris2022evaluating}.
Despite these efforts, integrating LMs with physics simulators for explanation generation remains an open problem. Recent works~\citep{memery2024simlmlanguagemodelsinfer, cherian2024llmphycomplexphysicalreasoning} highlight limitations of LMs in interpreting simulation traces.

%
\section{Method}

\subsection{Illustrative inverse problem involving a simulator and iterative optimization}
\label{subsec:illustrative-optimization-problem}

We illustrate our method using an optimization problem involving the design of a trajectory $\tau_\theta$ in $\mathbb{R}^2$ with parameters $\theta$ specifying the curve. Consider $\tau_\theta$ to also be parameterized by $u\in[0,1]$, so that $\tau_\theta(u=0)$ and $\tau_\theta(u=1) $ are the start and end points of the curve respectively.
We seek to minimize 
\begin{align}
    \label{eq:intobj}
    \argmin_\theta \left[ \; \DefInt {0} {1} {t(u)} {u} + \DefInt {0} {1} {c(u)} {u}  \;  \right]
\end{align} 
where the integrals correspond to the time taken for an object (say a train) to travel from $\tau_\theta(u=0)$ to $\tau_\theta(u=1) $ and the construction cost (say of the railroad) respectively. The functions $t(u)\in\mathbb{R}^+$ and  $c(u) \in \mathbb R^+$ are the time and cost associated with an infinitesimal length along the trajectory at $u$. The resulting optimization is non-trivial when the trajectory is constrained to negotiate obstacles (buildings, trees, ponds, etc.), since they affect dynamics (penalty on acceleration) as well as construction costs. 

We consider three forces acting on the train--driving force, frictional force and force due to air resistance--which lead to three components of the acceleration $a_d(u)$, $a_\mu(u)$ and $a_\eta(u)$ respectively. Obstacles impose their individual penalties on acceleration $a_o(u)$ leading to a net acceleration of $a = a_d + a_\mu + a_\eta + a_o$ where $a_d$ is proportional to the radius of curvature $\kappa(u)$, $a_\mu$ and $a_\eta$ depend on the velocity $v(u)$ and $a_o$ depends on the obstacle intersecting $\tau_\theta(u)$. 

We represent $\tau_\theta(u)$ using a \BC\ of order $n$ where $\theta$ represents the control points $\CPi i \in \mathbb{R}^2, \; i=0, 1, \cdots, n$. The trajectory \T\ and its derivative \Td\ (w.r.t. u) are given by
\begin{align}
    \nonumber \T = \sum\limits_{i=0}^n \C n i \; (1-u)^{n-i} \; u^i \; \CPi i  \quad \mathrm{and} \quad
    \Td = n \sum\limits_{i=0}^{n-1} \C {n-1} i  \; (1-u)^{n-1-i} \; u^i \; (\CPi {i+1} - \CPi i)
\end{align}
respectively while the curvature is $\kappa(u) = ||\Td \times \Tdd  || / || \Td ||^3 $. 
We discretize the integrals in Equation~\ref{eq:intobj} and approximate them as a left Riemann sum using $N$ intervals. 
The simulator proceeds in $N$ steps where at the $m^{th}$ step, the length of curved section $s_m$ is obtained in closed form and the net acceleration for the $m^{th}$ interval $a_m$ is calculated based on the current dynamics state and obstacle constraints. 
Using these, the time to travel the $m^{th}$ interval is obtained from elementary equations of motion as $t_m  = \left (-v_m + \sqrt{v_m^2 + 2a_m s_m} \right ) /a_m$. 
The cost $c_m$ for the $m^{th}$ interval is calculated as $c_m = \sigma_g(\mathbf x_m) + \sigma_o(\mathbf x_m)$ where $\mathbf x_m = \tau_\theta(m/N)$, $\sigma_g$ is a global cost associated with the terrain and $\sigma_o$ is a local cost associated with nearby obstacles. 
We use a Perlin noise model for the global cost. 
Finally, the velocity $v_{m+1}$ is updated via (explicit) time integration of the calculated acceleration and the calculation is repeated for the next step ($m+1$). 
The resulting optimization is
\begin{align}
    \label{eq:intobj}
    \argmin_\theta \sum\limits_{m=0}^{N} \left[ \;  \frac {-v_m + \sqrt{v_m^2 + 2a_m s_m} } {a_m}  \; + \; \left (\sigma_g(\mathbf x_m) + \sigma_o(\mathbf x_m) \right) s_m     \right]
\end{align} 
where all variables with the subscript $m$ depend on the control points, and hence on parameters $\theta$. 
At each iteration in the optimization, the simulation in Algorithm~\ref{alg:simstep} is run for $m$ ranging from $0$ to $N-1$ and the loss in the above optimization is calculated. 
The gradients (w.r.t. $\theta$) of the loss are calculated via autodifferentiation of $t_m$  and  $\sigma_o$.
\begin{algorithm}
    \caption{   \label{alg:simstep} A single iteration of the simulation.}
     \KwIn{ $m$, $v_m$, $\theta = \{\CPi i\}_{i=1,\cdots,n}$ \tcp*[r]{{step number, velocity, control points } }
        }
    \KwOut{$t_m$, $c_m$, $v_{m+1}$ \tcp*[r]{time and cost for $m^{th}$ interval, velocity for next step }} 
\vspace{1em}
    $\mathbf x_m \gets \tau_{\theta}(m/N)$\;
    $a_d \gets a \exp(-0.15 \kappa) $\tcp*[r]{driving acceleration}
    $a_\mu \gets -\mu_\mathrm{fric} \mathrm{sign}(v_m)$\tcp*[r]{deceleration due to frictional forces}
    $a_\eta \gets -\mu_\mathrm{air} v_m |v_m| $ \tcp*[r]{deceleration due to air resistance}
    $a_o \gets \mathrm{ObstacleConstraints}(\mathbf x_m)$\tcp*[r]{obstacle-induced deceleration}
    $a_m \gets a_d + a_mu + a_eta + a_o$ \;
    $s_m \gets \mathrm{ObjectConstraints}(m)$ \tcp*[r]{length of curve in $m^{th}$ interval}
    $t_m  \gets \left (-v_m + \sqrt{v_m^2 + 2a_m s_m} \right ) /a_m$ \tcp*[r]{time to traverse $m^{th}$ interval} 
    $c_m \gets \sigma_g(\mathbf x_m) + \sigma_o(\mathbf x_m)$ \tcp*[r]{construction cost for $m^{th}$ interval} 
    $v_{m+1} \gets  \text{clip}(v_m + a_m t_m, v_{\min}, v_{\max})$ \tcp*[r]{update velocity}
             \Return $(t_m, c_m, v_{m+1})$\;
\end{algorithm}

\subsection{Generating an optimization trace in natural language}
\label{subsec:explainable-optimization}

Our methodology consists of three phases: Emission of signals within the optimizer, functions to generate a trace in natural language, and  post hoc generation of explanations. 
At each step of the optimization process $k$, we instrument the optimizer to emit events, rewards, and parameter updates:
\begin{align}
\mathcal{E}_k &= \{e_j^i\}_{j=1}^{|E_k|} & & \text{a set of events: observable behavior of the system}\\
\mathcal{R}_k &= [r_k^1, r_k^2, \ldots] & & \text{a set of rewards: quantitative performance metrics}\\
\mathcal{U}_k &= \{\Delta \theta_j^k\}_{j=1}^p & & \text{a set of updates: changes to optimization parameters}
\end{align}
%
Then, we deploy transformation functions, ($\Phi_E$, $\Phi_R$, and $\Phi_U$) defined by the domain expert, to convert the signals emitted by the optimizer into natural language representations. 
In our illustrative example, $\Phi_E$ transforms a restricted set of event types: grid position (world coordinates translated to be within a $100 \times 100$ grid), speed, acceleration, curvature, object influence entry, and object influence exit. 
Each value is transformed to a simple natural language description that characterizes how the value changed since the last event step. 
For instance, acceleration changes are described as ``Change in acceleration: small (positive), at grid position (10, 20)'' where the magnitude (small, moderate, large, etc.) is scaled relative to the average change observed throughout the simulation, and the direction is indicated in parentheses.

$\Phi_R$ transforms reward information by capturing differences in time and cost since the last optimization step, i.e the difference between $R_k$ and $R_{k-1}$. These differences are scaled by the average change across the optimization process and labeled with qualitative descriptors, such as ``Change in time: high (positive)'' or ``Change in cost: moderate (negative).''

$\Phi_U$ transforms the change in control points since the last optimization step. For each control point, it characterizes the change in world coordinates, scaled by the average change across the entire optimization process. The natural language representation includes both magnitude and direction information, such as ``Control point 1: Magnitude: very small (ESE),'' where ESE indicates the east-southeast direction of movement. Importantly, $\Phi_U$ also incorporates contextual information about nearby obstacles, comparing previous and new positions: ``Previous position: small sandbox is close by, small valley is close by. New position: small sandbox is close by, small valley is close by.''


\subsection{Generating post hoc explanations from natural language traces}
\label{subsec:post-hoc-explanation-generation}

We then utilize a language model to generate descriptions of the optimization by conditioning on the natural language simulation trace. The description generation is a two step process:

\begin{enumerate}[leftmargin=1.2em, topsep=0em]
    \item \emph{Step-Level Descriptions}. For each step $k$, we generate a description utilising: preceding events $\Phi_E(\mathcal{E}_{k-1})$, current events $\Phi_E(\mathcal{E}_k)$, and parameter updates $\Phi_U(\mathcal{U}_k)$. 

    \item \emph{Global Description Synthesis} combines all step-level descriptions along with reward change information into a coherent narrative. Importantly, reward changes $\Phi_R(\mathcal{R}_k - \mathcal{R}_{k-1})$ are incorporated directly into this global synthesis. 
\end{enumerate} 

Importantly, event and update sample rates are utilized to control the granularity of the descriptions. An event sample rate of $10$ logs events every $10$ simulation steps, while an update sample rate of $5$ logs updates every $5$ simulation steps. This is useful for controlling the conciseness of descriptions.  

\pagebreak

\begin{figure*}[htbp]
	\centering
	\begin{subfigure}{0.48\textwidth}
		\centering
		\includegraphics[width=\textwidth]{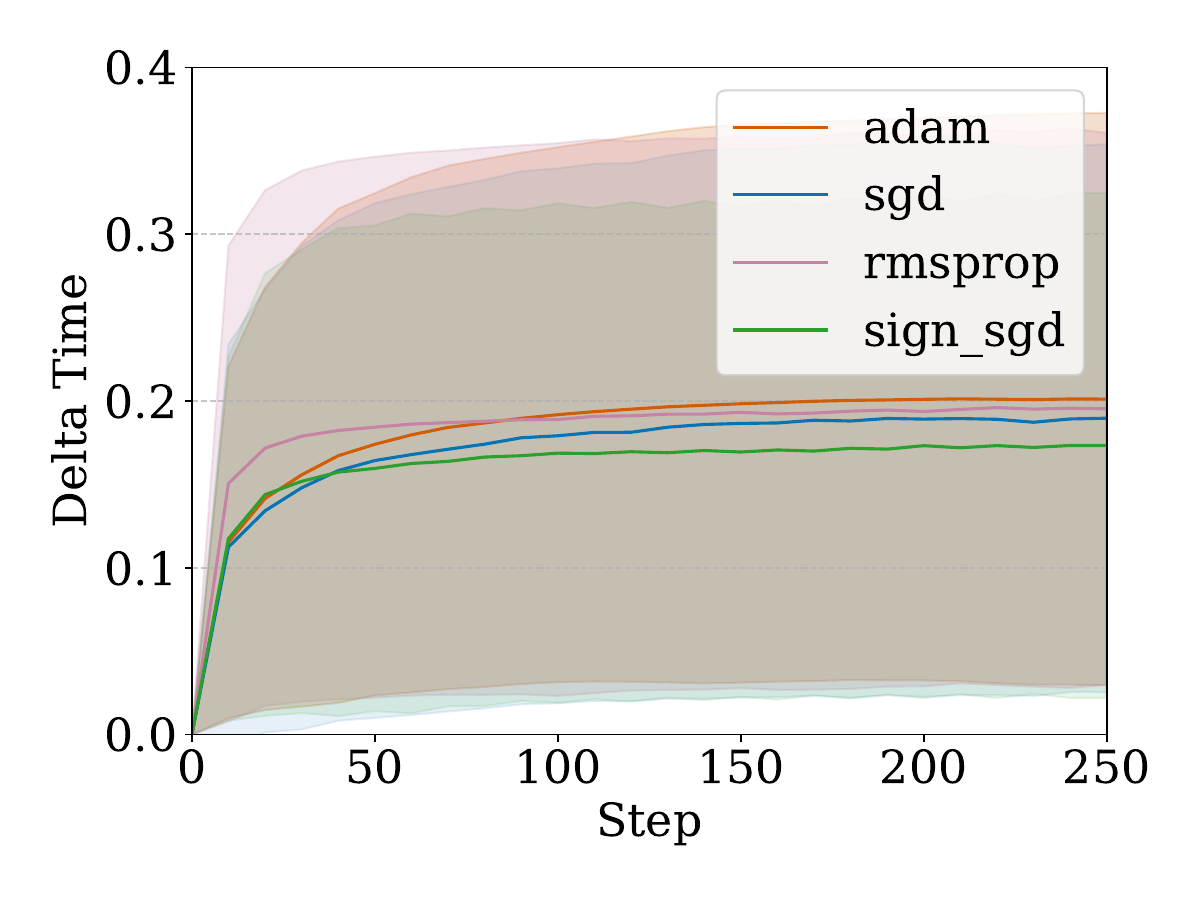}
		\label{fig:exp1_time}
	\end{subfigure}
	\hfill
	\begin{subfigure}{0.48\textwidth}
		\centering
		\includegraphics[width=\textwidth]{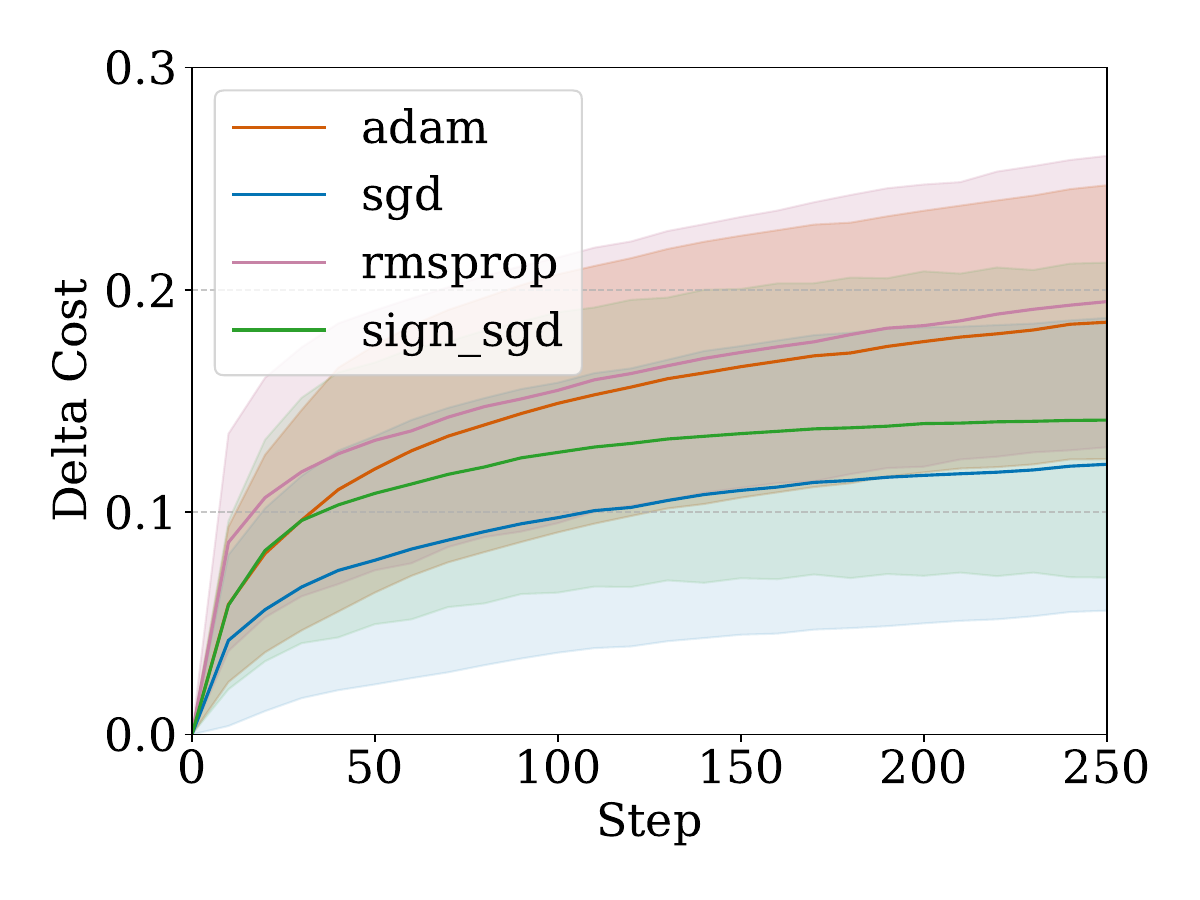}
		\label{fig:exp1_cost}
	\end{subfigure}
	\caption{Comparing relative savings in time (a) and cost (b) for different optimizers. }
	\label{fig:optimization_progress}
\end{figure*}

\section{Experimental Results}
\label{sec:experiments}
\subsection{Optimization Validation}
\label{subsec:optimization-validation}

We performed an extensive number of optimizations, across instances, using Adam, SGD, rmsprop, and sign\_sgd. We tested with linear, quadratic, and cubic objective functions. We evaluate them on $200$ random seeds, with $250$ optimization steps and the following parameters: learning rate 5e-3, number of obstacles $20$, number of control points $16$, and learning rate scheduler `cosine'. The results are presented in Figure~\ref{fig:optimization_progress}, as the relative savings for both time and cost, aggregated across random seeds to normalize for variability across instances (different obstacles and positions). 
As shown, both metrics are improved by the optimizers. Notably, the cost metric exhibits a larger absolute improvement compared to time, reflecting the optimization's effectiveness on this aspect of the objective. However, we observe substantial standard deviation across instances highlighting the suitability of our chosen optimization problem for robust and generalized evaluation. Further results are presented in the supplemental material. 

\subsection{Conversational Optimization Interaction}
\label{subsec:convo-opt}
We highlight the capabilities of our methodology by integrating our trace directly into an LM agent (GPT-4.1 for this section). 
This allows a natural interaction between the user and the domain expert, who can request certain optimizations and explanation in natural language. We exploit several key capabilities of LMs. First, the user can request a specific optimization to be run, having the LM initiating it and returning a detailed description of the process. 
Second, users can request specific state changes, and the LM calls upon tools to directly modify the state, e.g. adding or removing obstacles, moving control points, changing obstacle properties, etc. 
Finally, the LM requests to observe the output of $\Phi_E$ and $\Phi_U$ for a specific optimization directly. Our methodology enables examples, such as the following, for a user interacting with a domain-aware AI assistant. 
\paragraph{Example 1: Speed vs. Cost } Conversational exploration of optimizing for different priorities
\begin{coloreddialogue}
	\begin{dialogue}
		\small
		
		\speak{User} Run two optimizations: one prioritizing speed and one prioritizing cost.
		
		\speak{Assistant} ... speed-prioritized optimization: The system \textbf{rerouted the track to minimize slow-downs} ... skillfully avoiding obstacles like the "small statue," "small shed," and strategically passing closer to others (like the "small bush") when this meant a faster overall journey. Final adjustments involved "budgeting" minimal cost increases if it helped \textbf{maintain consistently high speed}. \textbf{Result: Time = 0.648s, Cost = 0.653} ... cost-prioritized optimization: Here, the optimizer's main objective was to \textbf{avoid expensive construction and penalty zones}. It aggressively rerouted to bypass dense obstacle clusters, even \textbf{tolerating modest speed losses if that kept costs down} ... \textbf{Result: Time = 0.712s, Cost = 0.612}. Summary: When you optimize for speed, the route is shaped for minimal slow-downs at the expense of some additional cost and potential length. When you optimize for cost, the path stays farther from expensive areas, sometimes sacrificing speed.
		
		\speak{User} Analyze and explain why the speed-prioritized optimization is faster and the cost-prioritized optimization is cheaper.
		
		\speak{Assistant} The speed-prioritized optimization is faster because it deliberately \textbf{allows the track to enter higher-cost zones} and maintains \textbf{smooth, high-speed arcs with limited braking} throughout the journey. You can see this from the many "large" positive speed changes in early and mid segments... Meanwhile, the cost-prioritized optimization is cheaper because it \textbf{steers through lower-cost portions of the grid}, often accepting higher curvature and extra acceleration/deceleration. It \textbf{avoids the cost jumps} seen in the speed-prioritized run by seeking cheaper construction ground. Specific examples show how the speed-optimized path uses smooth trajectories regardless of cost, while the cost-optimized path makes \textbf{strategic deviations around expensive areas—trading some speed for significant cost savings}.
	\end{dialogue}
\end{coloreddialogue}

\paragraph{Example 2: Obstacle Impact Analysis} In this second conversation, the user explores the impact of individual obstacles on optimization performance:
\begin{coloreddialogue}
	\begin{dialogue}
		\small
		\speak{User} Analyze the current state and identify which single obstacle would be most beneficial to remove to improve performance.
		
		\speak{Assistant} I've started systematically removing each obstacle one by one and running optimizations for each modified state... after removing the "small lamp" at [54, 86] and rerunning the optimization, I observed that the optimizer \textbf{exploited the new routing freedom to dramatically reduce both traversal time and construction cost}. The train navigated a \textbf{more direct path}, compressed penalties into sharp but brief events, and \textbf{focused the remainder of the optimization on fine-tuning}, all of which were not achievable with the lamp present.
	\end{dialogue}
\end{coloreddialogue}

\subsection{Question-answering Evaluation}
\label{subsec:qa-eval}

We evaluate our explanations by testing if they help an LM identify which obstacle removal would most improve performance. Using $50$ random seeds with $12$ obstacles and $12$ control points each, we asked the LM to predict obstacle removals that would maximize improvement in either time or cost metrics, for both original and optimized paths. Ground truth answers were determined by simulating each individual obstacle removal for original path questions, and by re-running optimizations for each removed obstacle for optimized path questions.

\begin{figure*}[htbp]
	\centering
	\includegraphics[width=.89\textwidth]{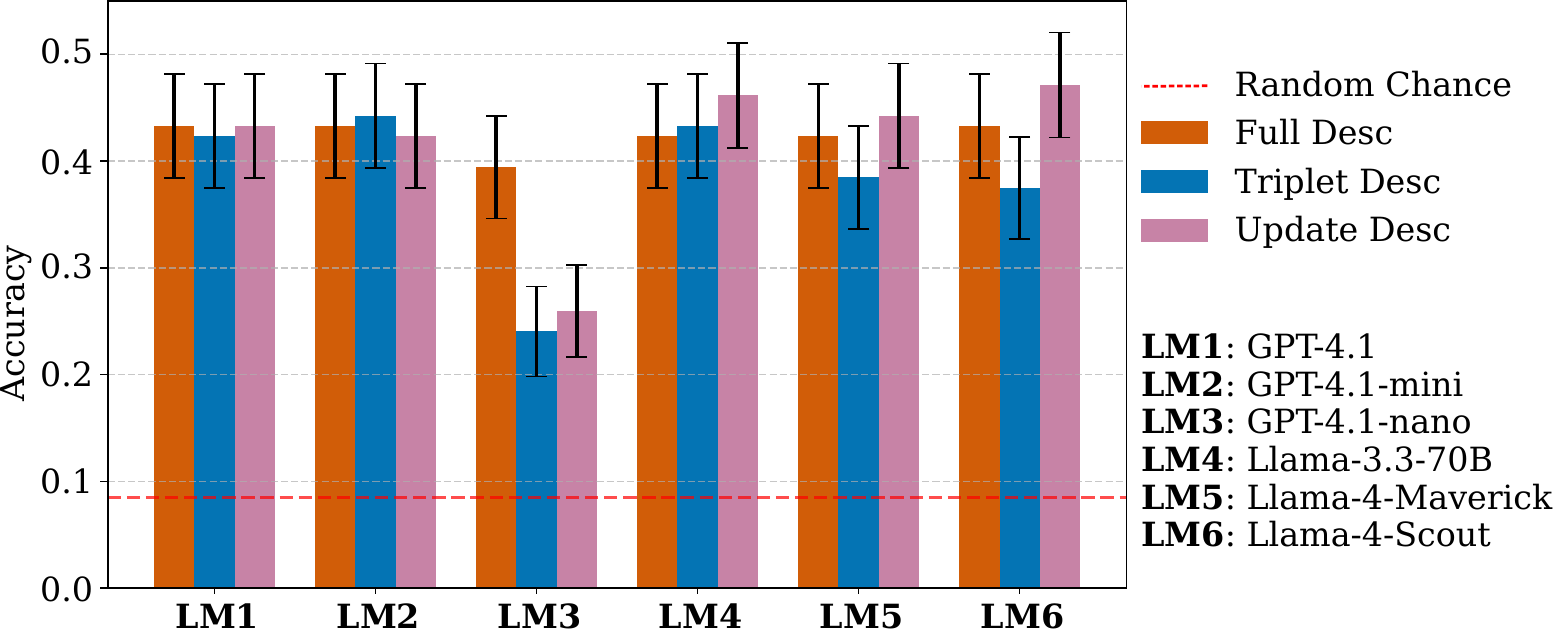}
	\caption{\label{fig:qa_results}  Question-answering well over chance. Different LMs were provided one of three descriptions and asked to identify the key obstacle to improving the performance of the path.}
\end{figure*}

The LM receives obstacle positions and radii, control point positions, and initial path events from $\Phi_E$, along with one of three explanation types: the complete optimization explanation (\textbf{Full description}), descriptions for each optimization step (\textbf{Step Level descriptions}), or the natural language updates from $\Phi_U$ for each step (\textbf{Update description}). Results from the question-answering experiment are shown in Figure~\ref{fig:qa_results}. 
All LMs evaluated outperform random chance (1/num\_obstacles = 0.085), indicating that the natural language descriptions produced by our method provide useful, actionable information. 
Performance is consistent across question types and description formats, with smaller Llama models~\citep{llama3modelcard,llama4modelcard} matching or even surpassing GPT-4.1. 
Notably, GPT-4.1-nano performs worse than expected, likely due to the greater context length required for step level and update descriptions, leading to less understanding in smaller models. 
Importantly, the full description achieves comparable accuracy to the longer, less summarized alternatives, demonstrating the efficiency and effectiveness of our approach for supporting downstream reasoning tasks. 

\subsection{Evaluation of robustness of explanations}
\label{subsec:lm-reconstruction-eval}


We address the evaluation of the question \textbf{``How informative are explanations?''}
by determining whether the parameters (optimized control points) can be discriminated from their noisy versions (distractors) when conditioning on the explanations of the whole optimization process and the control points' initial positions.
We report success rates in matching descriptions across different levels of noise using an LM. 
Our prompts, detailed in the supplementary material, provide the grid positions (within a $100 \times 100$ grid--hereafter referred to as grid coordinate space) of the initial control points rather than the exact or floating-point numerical positions, in order to alleviate LMs' susceptibility to numerical understanding and processing abilities (NUPA - ~\cite{yang2025number}).

We generate explanations via three types of descriptions as in Section~\ref{subsec:qa-eval}: Full, Step Level, and Update.
We generate distractors by adding Gaussian noise to the target control points' final positions with increasing standard deviations $\sigma\in\{0.02,0.06,0.1,0.15,0.2\}$, within the floating-point coordinate space. We expect that the larger the value of $\sigma$ the easier the discrimination task will be.
To keep a relatively small context length, we experiment with one distractor for each task. The LMs are prompted with $2$ candidates (in an unknown, random order), one is a distractor while the other is the target that must be identified. As a control (\textit{Numerical}), we use a method where the LM knows grid positions directly (without using the explanations). This represents a `best case' scenario and could be considered an upper bound. 

We experiment with a dataset of $10$ randomly sampled instances (obstacles, cost Perlin noise, initial control points) and hyperparameters detailed in the supplementary material. 
We aggregate results for each method over this dataset with $2$ seeds each for distractor generation and $2$ seeds to condition the LM test. This yields $40$ data points per method. 

\begin{figure*}
	\centering
	\includegraphics[width=.8\textwidth]{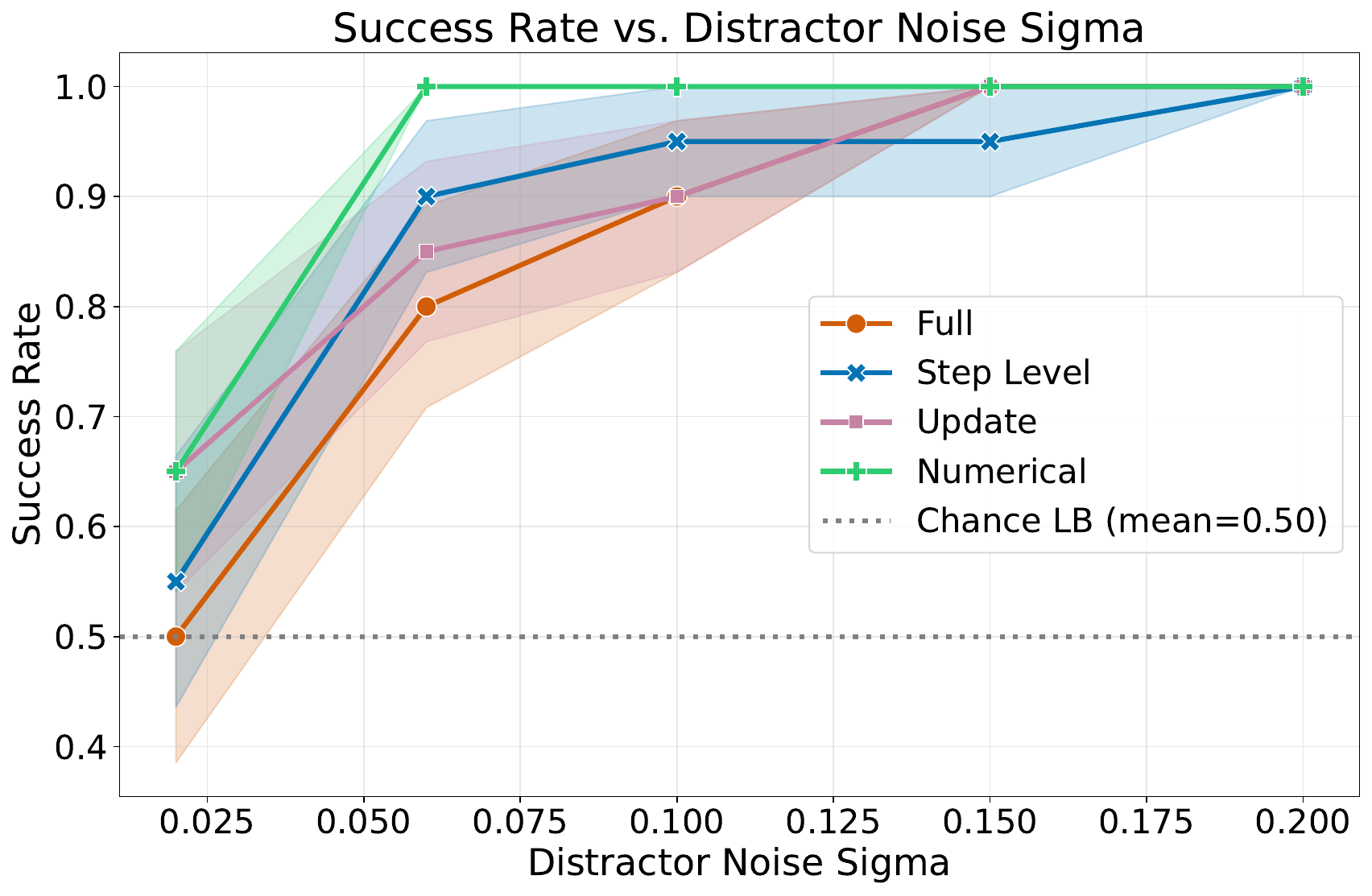}
	\caption{
		\label{fig:lm-discrimination-results} 
		Success rates of Llama-4-Maverick\citep{llama4modelcard} at the discrimination tasks, for different values of the standard deviation $\sigma$ parameterising the distractor generation.
	}
\end{figure*}

The results are plotted in Figure~\ref{fig:lm-discrimination-results}, showing that, even at the lowest noise level ($\sigma= 0.02$), most explanation types perform above chance (0.50), demonstrating that LMs can extract meaningful information from our explanations of optimization processes. 
The quick convergence (as soon as $\sigma\geq0.06$) of all methods to near-perfect performance as distractor noise levels increase validates that our explanations are informative to some extent.
On the other hand, the performance differences at lower noise levels highlight the importance of the explanation design when subtle discrimination is required.
We leave it to subsequent work to consider more sophisticated explanation designs.
We refer readers to the supplementary material for results using different LMs.

\begin{figure*}[htbp]
	\centering
	\includegraphics[width=\textwidth]{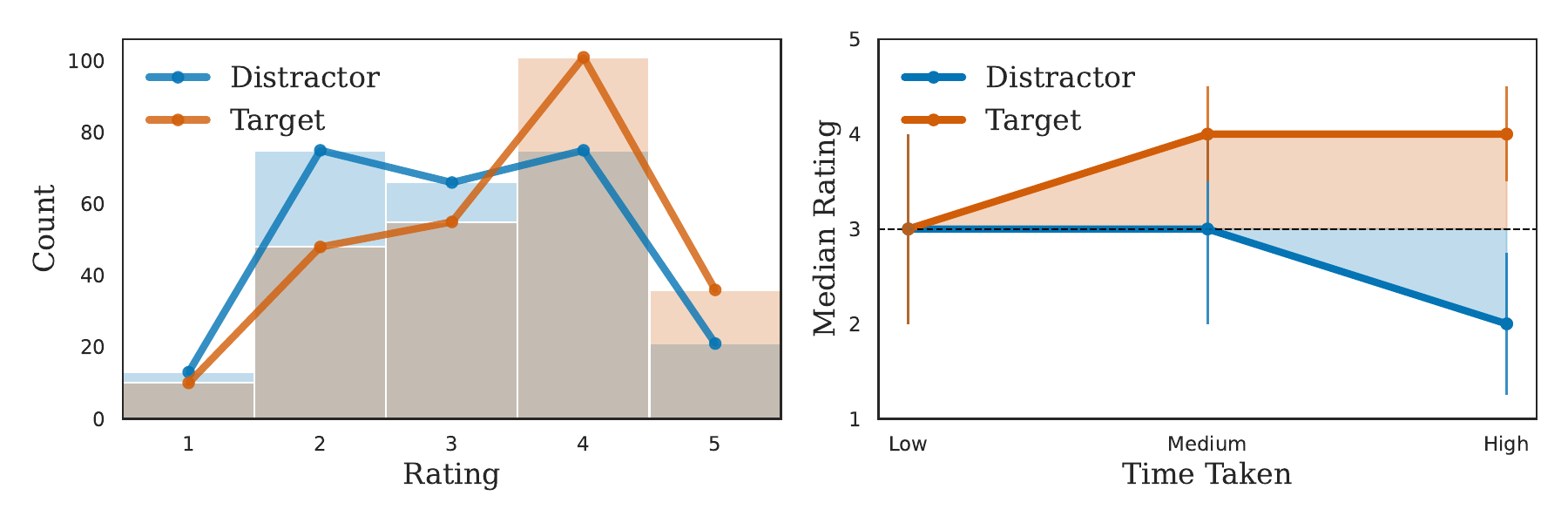}
	\caption{\label{fig:user_study_results} Our user study. Left: a histogram of the ratings for our explanations against the correct (target) and a distractor. Right: the median rating vs time taken by respondents.}
\end{figure*}

\subsection{Evaluation of explanations via survey}
\label{subsec:user-study}

To evaluate our generated explanations, we conducted a user study with $50$ participants on Prolific. A detailed explanation of the optimization task and survey questions (see supplemental material) was provided and each participant rated how well a text description (generated by our method) matched a visualization, using a $1$--$5$ scale. Each visualization showed five static images from intermediate stages of an optimization, illustrating the change in control points at each step. Participants rated five descriptions, presented in random order.

For each description, two sets of images were presented. The optimization matching the description (referred to as the target) and a different set where obstacle positions were shuffled (referred to as the distractor). These sets were presented in random order, and participants were asked to rate how well the description matched each variant. The time taken for participants to complete each rating was measured to assess decision confidence and difficulty. Each participant was paid at least the UK minimum hourly wage, with an average survey time of $12$ minutes.

The results for the user study are shown in Figure~\ref{fig:user_study_results}. The total count of ratings for each variant of the optimization (left) shows a statistically significant preference for the target variant over the distractor variant. The median rating for each variant is plotted grouped by the time taken for the response. The time is quantized into bins: Low' refers to the lowest third of the survey time, `Medium' refers to the middle third of the survey time, and `High' refers to the highest third of the survey time. Error bars visualize the interquartile range of the ratings and time taken. The plot shows a clear trend of higher ratings for the target variant by respondents who took longer to complete the survey. The inverse is observed for the distractor variant.

\subsection{Application to a different optimization problem: Explainable LM training}
\label{sec:lm_training}

    


To demonstrate the generality of our methodology, we apply it to the training of a small LM on a physics question-answering dataset \footnote{Both the model and code are adapted from \url{https://github.com/karpathy/nanoGPT}}. This experiment aims to illustrate that our framework is not limited to physics optimization, but can also be extended to complex neural network training processes. Specifically, we wish to explore creating explanations about a system where the relationship between parameters and model behavior is black-box and unclear to the domain expert. 

During each validation step $k$ of training, we capture three key components: (i) \textbf{Events} ($E_k$), consisting of 10 generated text samples; (ii) \textbf{Rewards} ($R_k$), represented by the training and validation losses; and (iii) \textbf{Updates} ($U_k$), measured as the change in average parameter values per layer. $\Phi_E$ preserves the raw text samples as human-interpretable events, $\Phi_R$ normalizes loss differences and maps them to qualitative descriptors, and $\Phi_U$ combines the layer names with the relative magnitudes of change between steps and bins them into qualitative categories. We present qualitative examples of this process in the results section, for a more in depth analysis see supplemental material. 

We demonstrate our framework’s application to LM training, presenting both a global explanation of the training process and a stepwise explanation for a specific validation step. These natural language outputs are generated directly by our method, see supplemental material for further examples.

\begin{coloreddialogue}
	\textbf{Global description:} 
	\begin{dialogue} 
		\small \speak{Assistant}Over the course of training, the language model exhibited a \textbf{marked shift from incoherent, random outputs} to \textbf{increasingly structured, contextually appropriate responses} to scientific prompts. Early optimization steps were dominated by \textbf{large-scale changes in embedding and attention parameters}, which rapidly transitioned the model from symbol-level noise to the production of basic scientific terminology and syntax. As training progressed, \textbf{refinements in LayerNorm and attention submodules} became the primary drivers of improvement, leading to \textbf{enhanced grammaticality, topical relevance}, and a \textbf{reduction in repetitive or degenerate outputs}. Notably, the most significant qualitative gains were linked to \textbf{targeted updates in attention and normalization layers}, underscoring their role in enabling the model to integrate context and manage information flow. 
	\end{dialogue}
\end{coloreddialogue}

\begin{coloreddialogue}
	\textbf{Stepwise example (Validation Step $k$):}
	\begin{dialogue} 
		\small 
		\speak{Assistant}
		At validation step $k$, a targeted update to the \textbf{LayerNorm and attention output projection weights} resulted in a \textbf{notable improvement in the model's generated answer}. Prior to this step, the model's output was: \ \texttt{"The force is... qwerty zxv mass..."} \ Following the update, the generated text became: \ \texttt{"The force on an object is equal to its mass times acceleration, according to Newton's second law."} \ This qualitative leap was directly attributable to \textbf{improved contextual integration and normalization}, as highlighted by our framework's attribution maps. The loss decreased only marginally, but the \textbf{semantic and syntactic coherence of the output increased substantially}, demonstrating how parameter changes in specific submodules can drive interpretable, fine-grained improvements in model behavior.
	\end{dialogue}
\end{coloreddialogue}

Thus, our methodology enables interpretable explanations that relate specific parameter updates to observable changes in model outputs. The global description summarizes the overall trajectory of learning, while the stepwise example demonstrates the framework’s ability to attribute qualitative improvements to targeted changes in model components, even when the loss shows minor differences.

\section{Discussion and conclusion}
\label{subsec:discussion}


\textbf{LMs comprehend and interact with optimization processes}: 
Section \ref{subsec:optimization-validation} demonstrates how LMs successfully interact with optimization systems through our generated explanations. These explanations serve as an effective interface between the LM and system, enabling sophisticated domain-specific interactions. This approach is highly modular and extensible, limited only by LM capabilities and user-defined events, updates, and rewards.

\textbf{Method flexibility across domains}: 
We demonstrate our method in both physics optimization and language model training, representing white-box and black-box systems respectively. This flexibility addresses the challenge of explaining optimization processes where parameter-behavior relationships may be clear but the rationale for parametric changes remains opaque. Our framework accommodates both system types through user-configurable emission and transformation functions.

\textbf{Explanations enable domain-specific decision making}:
Section~\ref{subsec:qa-eval} shows that for tasks like obstacle identification, LMs can successfully leverage explanations to identify high-impact elements. Performance varies by task type—spatial reasoning tasks show promising results while coordinate estimation proves more challenging without explicit numerical information. Thus, explanation utility depends on both task nature and information representation. With human evaluation, in Section~\ref{subsec:user-study}, we see a clear preference for the target over the distractor, with a median rating of $4/5$ for the target and $2/5$ for the distractor when users are within the top 33\% of survey time. We believe this shows the usefulness of our explanations for users to identify the corresponding optimization, when taking the time to fully interpret the explanations.

\subsection{Limitations}
\label{subsec:limitations}


We observe diminishing returns with scale in optimization complexity (e.g. number of obstacles and control points) partly due to difficulties in evaluating explanations. Our method depends on user-defined events, rewards, and updates, which might be challenging in some domains.  Finally, although explainability can largely be considered as positive and impactful, there are some negative societal impacts that have been theorized~\cite{xaipitfallsdarkpatterns}.

\subsection{Conclusion and Future Work}

We demonstrated an effective methodology for enabling natural language explanations of optimization processes for inverse problems with simulators.  We meld numerical optimization and human interpretability, and demonstrate that explanations effectively convey optimization behaviors across domains, with improvements in downstream tasks.
Future work could: (1) handle scaling to more complex optimization scenarios through improved explanation strategies; (2) develop interactive interfaces enabling users to interact through natural language with more complex optimization tasks; and (3) integrate with other XAI techniques in domains like robotics and autonomous systems where understanding optimization decisions is critical for safety and trust.

\begin{ack}
This work was funded by the UKRI Centre for Doctoral Training in Natural Language Processing (grant EP/S022481/1) and ELIAI (The Edinburgh Laboratory for Integrated Artificial Intelligence) EPSRC (grant EP/W002876/1).
\end{ack}

\bibliographystyle{plain}
\bibliography{main}

\newpage

\begin{center}
    {\LARGE \textbf{Supplemental Material}}
\end{center}

\section*{Contents}

This supplementary material provides detailed resources and supporting information for the main paper. Additionally, there are examples in the following HTML files (\textbf{NOTE}: hyperlinks may not open in some PDF readers, please open the HTML files in a web browser):
\begin{itemize}
    \item \url{full_example_1.html}
    \item \url{full_example_2.html}
    \item \url{exp1_example_1.html} (with full output logs at \url{logs/exp1_example_1.json})
    \item \url{exp1_example_2.html} (with full output logs at \url{logs/exp1_example_2.json})
    \item \url{exp1_example_3.html} (with full output logs at \url{logs/exp1_example_3.json})
\end{itemize}
Specifically, this document includes:

\begin{enumerate}
    \item \textbf{Exact outputs of the physics optimization environment:} Step-by-step logs, event traces, and simulation results from the differentiable trajectory optimization experiments described in the main text (see Section~\ref{app:physics-env}).
    \item \textbf{Language model prompts:} Full prompt templates and example inputs for:
    \begin{itemize}
        \item DSPy pipeline templates (see Section~\ref{app:lm-prompts:template})
        \item Explanation generation (see Section~\ref{app:lm-prompts:explanation-generation})
        \item Conversational agent experiment (see Section~\ref{app:lm-prompts:conversational-interaction})
        \item Question-answering experiment (see Section~\ref{app:lm-prompts:qa-eval})
        \item Language model discrimination experiment (see Section~\ref{app:lm-prompts:lm-discrimination-eval})
        \item Explainable language model training and explanation generation (see Section~\ref{app:lm-prompts:explainable-lm-training})
    \end{itemize}
    \item \textbf{Further examples:} Additional outputs and qualitative examples for:
    \begin{itemize}
        \item Optimization validation (see Section~\ref{app:results:optimization-validation})
        \item Conversation examples (see Section~\ref{app:results:conversational-interaction})
        \item Question answering (see Section~\ref{app:results:question-answering-evaluation})
        \item Language model discrimination (see Section~\ref{app:results:lm-discrimination-eval})
        \item Explainable language model training (see Section~\ref{app:results:explainable-lm-training})
    \end{itemize}
    \item \textbf{Further information on the user study:} Expanded methodology, survey instruments, and analysis details for the human evaluation component (see Section~\ref{app:user-study}).
\end{enumerate}

\pagebreak
\section*{Further Details on Physics Optimization Environment}
\label{app:physics-env}

This section provides additional details about our physics-based optimization environment that serves as the illustrative example throughout the paper. See the additional HTML for visualizations of the optimization process and the output of the system in terms of events, rewards, and updates, found at:
\begin{itemize}
    \item \href{./full_example_1.html}{HTML Example 1} (file \url{full_example_1.html})
    \item \href{./full_example_2.html}{HTML Example 2} (file \url{full_example_2.html})
\end{itemize}

The table below, Table \ref{tab:simulation-vars}, shows the evolution of key simulation variables during the first 25 steps (out of 100) of a randomly initialized optimization instance. 

\begin{table}[htbp]
	\centering
	\begin{tabular}{c|r|r|r|r|r|r}
	\textbf{Step} & \textbf{time} & \textbf{acceleration} & \textbf{air resistance} & \textbf{cost} & \textbf{curvature} & \textbf{velocity} \\
	\hline
	1  & 0.0175 & 4.2486 & 0.5000 & 0.3973 & 0.0689 & 1.0745 \\
	2  & 0.0347 & 4.2084 & 0.5773 & 0.3690 & 0.0191 & 1.1467 \\
	3  & 0.0515 & 4.1000 & 0.6575 & 0.3389 & 0.0569 & 1.2155 \\
	4  & 0.0679 & 4.0138 & 0.7387 & 0.3072 & 0.0637 & 1.2813 \\
	5  & 0.0839 & 3.9412 & 0.8208 & 0.2740 & 0.0509 & 1.3444 \\
	6  & 0.0995 & 3.8769 & 0.9037 & 0.2391 & 0.0260 & 1.4051 \\
	7  & 0.1149 & 3.8083 & 0.9872 & 0.2029 & 0.0060 & 1.4635 \\
	8  & 0.1299 & 3.6981 & 1.0710 & 0.2298 & 0.0414 & 1.5190 \\
	9  & 0.1446 & 3.5883 & 1.1538 & 0.2994 & 0.0778 & 1.5719 \\
	10 & 0.1590 & 3.4805 & 1.2354 & 0.3613 & 0.1131 & 1.6221 \\
	11 & 0.1732 & 3.3761 & 1.3156 & 0.4150 & 0.1460 & 1.6700 \\
	12 & 0.1872 & 3.2758 & 1.3944 & 0.4601 & 0.1753 & 1.7157 \\
	13 & 0.2009 & 3.1801 & 1.4717 & 0.4962 & 0.2005 & 1.7593 \\
	14 & 0.2144 & 3.0895 & 1.5475 & 0.5231 & 0.2209 & 1.8009 \\
	15 & 0.2277 & 3.0042 & 1.6217 & 0.5404 & 0.2363 & 1.8409 \\
	16 & 0.2407 & 2.9242 & 1.6944 & 0.5481 & 0.2464 & 1.8791 \\
	17 & 0.2537 & 2.8494 & 1.7656 & 0.5343 & 0.2514 & 1.9159 \\
	18 & 0.2664 & 2.7797 & 1.8354 & 0.5250 & 0.2513 & 1.9513 \\
	19 & 0.2789 & 2.7150 & 1.9037 & 0.5208 & 0.2463 & 1.9853 \\
	20 & 0.2913 & 2.6550 & 1.9707 & 0.5219 & 0.2366 & 2.0181 \\
	21 & 0.3035 & 2.5994 & 2.0364 & 0.5282 & 0.2226 & 2.0498 \\
	22 & 0.3155 & 2.5481 & 2.1008 & 0.5398 & 0.2046 & 2.0804 \\
	23 & 0.3273 & 2.5006 & 2.1640 & 0.5567 & 0.1829 & 2.1100 \\
	24 & 0.3390 & 2.4569 & 2.2261 & 0.5756 & 0.1580 & 2.1387 \\
	25 & 0.3505 & 2.4165 & 2.2869 & 0.5581 & 0.1300 & 2.1664 \\
	\end{tabular}
	\caption{Variables in the first 25 simulation steps (out of 100) of the train track optimization problem, where the Perlin noise, obstacles, and position of control points are sampled randomly.}
	\label{tab:simulation-vars}
\end{table}
	
\begin{table}[htbp]
	\centering
	\begin{tabular}{|c|p{11cm}|}
	\hline
	\textbf{Timesteps} & \textbf{Event Summary} \\
	\hline
	0--4 & 
	\begin{itemize}
		\item (5,5) $\rightarrow$ (10,10) Change in speed: very large (positive)
		\item (5,5) $\rightarrow$ (10,10) Change in acceleration: small (negative)
		\item (5,5) $\rightarrow$ (10,10) Change in cost: medium (negative)
		\item (5,5) $\rightarrow$ (10,10) Change in curvature: very small (negative)
		\item (5,5) $\rightarrow$ (10,10) Change in air resistance: very large (positive)
	\end{itemize}
	\\
	\hline
	5--9 & 
	\begin{itemize}
		\item (12,12) $\rightarrow$ (18,18) Change in speed: very large (positive)
		\item (12,12) $\rightarrow$ (18,18) Change in acceleration: small (negative)
		\item (12,12) $\rightarrow$ (18,18) Change in cost: medium (positive)
		\item (12,12) $\rightarrow$ (18,18) Change in curvature: very small (positive)
		\item (12,12) $\rightarrow$ (18,18) Change in air resistance: very large (positive)
	\end{itemize}
	\\
	\hline
	10--14 & 
	\begin{itemize}
		\item (20,19) $\rightarrow$ (26,26) Change in speed: very large (positive)
		\item (20,19) $\rightarrow$ (26,26) Change in acceleration: small (negative)
		\item (20,19) $\rightarrow$ (26,26) Change in cost: large (positive)
		\item (20,19) $\rightarrow$ (26,26) Change in curvature: very small (positive)
		\item (20,19) $\rightarrow$ (26,26) Change in air resistance: very large (positive)
	\end{itemize}
	\\
	\hline
	15--19 & 
	\begin{itemize}
		\item (28,28) $\rightarrow$ (35,35) Change in speed: very large (positive)
		\item (28,28) $\rightarrow$ (35,35) Change in acceleration: very small (negative)
		\item (28,28) $\rightarrow$ (35,35) Change in cost: small (negative)
		\item (28,28) $\rightarrow$ (35,35) Change in curvature: very small (negative)
		\item (28,28) $\rightarrow$ (35,35) Change in air resistance: very large (positive)
	\end{itemize}
	\\
	\hline
	20--24 & 
	\begin{itemize}
		\item (36,37) $\rightarrow$ (43,44) Change in speed: large (positive)
		\item (36,37) $\rightarrow$ (43,44) Change in acceleration: very small (negative)
		\item (36,37) $\rightarrow$ (43,44) Change in cost: small (positive)
		\item (36,37) $\rightarrow$ (43,44) Change in curvature: very small (negative)
		\item (36,37) $\rightarrow$ (43,44) Change in air resistance: very large (positive)
	\end{itemize}
	\\
	\hline
	\end{tabular}
	\caption{Summary of events for each batch of 5 timesteps.}
	\label{tab:events-mapping}
	\end{table}
	
The numerical values shown above are then mapped to the following natural language events that are used to generate descriptions of the optimization process. Examples of how the first 25 timesteps are mapped are shown in Table \ref{tab:events-mapping}.

The rewards of the simulation are then also mapped to natural language descriptions, utilising the previous steps reward values, this is shown below for the same optimization example.

\begin{tcolorbox}[colback=red!5!white, colframe=red!75!black, title=Reward Deltas]
	\begin{itemize}
		\item Change in time: very large (negative)
		\item Change in cost: very large (negative)
		\item Change in total: very large (negative)
		\item Change in length: very large (negative)
	\end{itemize}
\end{tcolorbox}

Finally, the update to the control points are also mapped to natural language descriptions, this is shown below for the same optimization example (six of ten control points are shown).

\noindent
\begin{minipage}[t]{0.48\textwidth}
\begin{tcolorbox}[title=Control Point 1, colback=gray!5!white, colframe=black]
\textbf{Old:} $(0.1708,\ 0.1688)$ \\
\textbf{New:} $(0.1701,\ 0.1682)$ \\[0.5em]
Magnitude: small (SW)

Previous position: small bench is close by, big building is close by.

New position: small bench is close by, big building is close by.
\end{tcolorbox}
\end{minipage}
\hfill
\begin{minipage}[t]{0.48\textwidth}
\begin{tcolorbox}[title=Control Point 2, colback=gray!5!white, colframe=black]
\textbf{Old:} $(0.3418,\ 0.3431)$ \\
\textbf{New:} $(0.3424,\ 0.3431)$ \\[0.5em]
Magnitude: very small (ENE)

Previous position: small bench is close by, big building is close by, small tower is close by.

New position: small bench is close by, big building is close by, small tower is close by.
\end{tcolorbox}
\end{minipage}

\vspace{0.5em}

\noindent
\begin{minipage}[t]{0.48\textwidth}
\begin{tcolorbox}[title=Control Point 3, colback=gray!5!white, colframe=black]
\textbf{Old:} $(0.5062,\ 0.4595)$ \\
\textbf{New:} $(0.5074,\ 0.4596)$ \\[0.5em]
Magnitude: small (ENE)

Previous position: small fountain is close by, big barrel is close by, small bush is close by, small tower is close by.

New position: small fountain is close by, big barrel is close by, small bush is close by, small tower is close by.
\end{tcolorbox}
\end{minipage}
\hfill
\begin{minipage}[t]{0.48\textwidth}
\begin{tcolorbox}[title=Control Point 4, colback=gray!5!white, colframe=black]
\textbf{Old:} $(0.6500,\ 0.7422)$ \\
\textbf{New:} $(0.6511,\ 0.7418)$ \\[0.5em]
Magnitude: very small (SE)

Previous position: big gate is close by, big barrel is close by.

New position: big gate is close by, big barrel is close by.
\end{tcolorbox}
\end{minipage}

\vspace{0.5em}

\noindent
\begin{minipage}[t]{0.48\textwidth}
\begin{tcolorbox}[title=Control Point 5, colback=gray!5!white, colframe=black]
\textbf{Old:} $(0.7024,\ 0.9027)$ \\
\textbf{New:} $(0.7020,\ 0.9024)$ \\[0.5em]
Magnitude: small (WSW)

Previous position: No objects are close by, nearest is big pond.

New position: No objects are close by, nearest is big pond.
\end{tcolorbox}
\end{minipage}
\hfill
\begin{minipage}[t]{0.48\textwidth}
\begin{tcolorbox}[title=Control Point 6, colback=gray!5!white, colframe=black]
\textbf{Old:} $(0.9983,\ 0.9783)$ \\
\textbf{New:} $(0.9982,\ 0.9782)$ \\[0.5em]
Magnitude: very small (SW)

Previous position: big pond is close by.

New position: big pond is close by.
\end{tcolorbox}
\end{minipage}

\pagebreak
\section*{Language Model Prompts \& Implementation Details}
\label{app:lm-prompts}


\subsection*{Template Prompt}
\label{app:lm-prompts:template}

We present a template prompt in Listing~\ref{lst:template} with interlocutors highlighted in blue with framing, and variables highlighted in green.
It corresponds mostly to the common template implemented by DSPy~\citep{khattab2024dspy} to reproduce (zero-shot) Chain-of-Thought prompting~\citep{wei2022CoTPromptingElicitsReasoning}, as seen in the \pgvar{CoT\_reasoning} variable.

\lstset{caption={Template Prompt}, label={lst:template}}
\begin{lstlisting}[escapechar=@]
@\pfVar{blue}{SYSTEM:}@
Your input fields are:
@\pgvar{list\_input\_fields}@

Your output fields are:
@\pgvar{CoT\_reasoning}@
@\pgvar{list\_output\_fields}@

All interactions will be structured in the following way, with the appropriate values filled in.

[[ ## @\pgvar{input\_field1.name}@ ## ]]
{@\pgvar{input\_field1.name}@}

@  $\cdots$ @
 
[[ ## @\pgvar{input\_fieldN.name}@ ## ]]
{@\pgvar{input\_fieldN.name}@}


[[ ## completed ## ]]

In adhering to this structure, your objective is: 
@ \pgvar{task\_description} @

@\pfVar{blue}{USER:}@
[[ ## @\pgvar{input\_field1.name}@ ## ]]
{@\pgvar{input\_field1.value}@}

@ $\cdots$ @
 
[[ ## @\pgvar{input\_fieldN.name}@ ## ]]
{@\pgvar{input\_fieldN.value}@}

Respond with the corresponding output fields, starting with the field `[[ ## @\pgvar{CoT\_reasoning}@ ## ]]`, then `[[ ## @\pgvar{output\_field1.name}@ ## ]]`, then @$\cdots$@, then `[[ ## @\pgvar{output\_fieldM.name}@ ## ]]`, and then ending with the marker for `[[ ## completed ## ]]`.


\end{lstlisting}

\subsection*{Description/Explanation Generation Prompts}
\label{app:lm-prompts:explanation-generation}

Listing~\ref{lst:descriptions:td} shows the LM prompt for generating descriptions/explanations for the \textit{Full} Description/Explanation Generation prompt.

\lstset{caption={\textit{Full} Description/Explanation Generation - Task Description}, label={lst:descriptions:td}}
\begin{lstlisting}[escapechar=@]
There is a simulation of a train moving along a track. The train wants to get to the end of the track as fast as possible and also for the track to cost as little as possible to construct. There are obstacles that slow the train down and each position on the track has a cost to construct. A differentiable simulation is used to optimise the track by updating control points that define where the track gets layed. A simulation is run where events are recorded, then the optimiser updates the control points of the track based on the time and cost of the track, then a new simulation is run with the events recorded.

You are presented with a description of the optimization process, i.e. a sequence of descriptions of update steps and the events that they caused. You are also shown the change in rewards that each update brought about. Combine the descriptions of each update with the change in rewards to create a global story of the optimization that makes precise reference to the change in obstacles, control points, events, updates, and rewards.

It is very important to note that the reader of this summary is already an expert in the field of optimization and machine learning, so the summary should be written in a way that is accessible to them and does not give general or simple information and should be very very detailed, with specific references to obstacles, control points, events, updates, and rewards.
\end{lstlisting}
The LM receives the following inputs:
\begin{itemize}
    \item \texttt{description}: A description of the simulation setup or context.
\end{itemize}
The LM is prompted to return:
\begin{itemize}
    \item \texttt{summary}: A detailed summary of the optimisation process.
\end{itemize}

Listing~\ref{lst:descriptions:step-level-td} shows the LM prompt for generating descriptions/explanations for the \textit{Step Level} Description/Explanation Generation prompt.

\lstset{caption={\textit{Step Level} Description/Explanation Generation - Task Description}, label={lst:descriptions:step-level-td}}
\begin{lstlisting}[escapechar=@]
There is a simulation of a train moving along a track. The train wants to get to the end of the track as fast as possible and also for the track to cost as little as possible to construct. There are obstacles that slow the train down and each position on the track has a cost to construct. A differentiable simulation is used to optimise the track by updating control points that define where the track gets layed. A simulation is run where events are recorded, then the optimiser updates the control points of the track based on the time and cost of the track, then a new simulation is run with the events recorded.

You are presented with a single optimization step of events before and after an update. Describe the relationship between the events and the update in great detail, i.e. describe in detail what the update has done to the events. Try your best to discover hidden relationships between the events and the update and be very specific.
    
It is very important to note that the reader of this summary is already an expert in the field of optimization and machine learning, so the summary should be written in a way that is accessible to them and does not give general or simple information and should be very very detailed, with specific references to obstacles, control points, events, updates, and rewards.
\end{lstlisting}

The LM receives the following inputs:
\begin{itemize}
    \item \texttt{events\_before}: List of events before the update.
    \item \texttt{update}: Update description.
    \item \texttt{events\_after}: List of events after the update.
\end{itemize}
The LM is prompted to return:
\begin{itemize}
    \item \texttt{explanation}: Explanation of how the update relates to the event changes.
\end{itemize}

\subsection*{Conversational Interaction Prompts}
\label{app:lm-prompts:conversational-interaction}

This section details the prompts used for the conversational interaction capabilities of the system. Two main language model (LM) signatures are employed: one for understanding and translating state modification requests, and another for the main conversational agent logic.

\subsubsection*{State Modification Prompt}

The first signature, \texttt{StateModifierSignature}, is designed for an LM tasked with converting natural language descriptions of desired simulation state changes into a formal list of state-changing commands. This leverages a Chain-of-Thought (CoT) \citep{wei2022CoTPromptingElicitsReasoning} process to ensure accurate translation. Listing~\ref{lst:state-modifier-signature} presents the task description provided to the LM for this purpose.

\lstset{caption={State Modifier Signature - Task Description}, label={lst:state-modifier-signature}}
\begin{lstlisting}[escapechar=@]
There is a simulation of a train moving along a track. The train wants to get to the end of the track as fast as possible and also for the track to cost as little as possible to construct. There are obstacles that slow the train down and each position on the track has a cost to construct. A differentiable simulation is used to optimise the track by updating control points that define where the track gets layed. 

Convert a natural language description of changes to a simulation state into a formal list of state-changing commands.

IMPORTANT: All positions are specified using a 100x100 grid system where [0,0] is the bottom-left corner and [99,99] is the top-right corner.

The supported commands are:
- add_obstacle: Add a new obstacle at a specified position with properties (center, radius, penalty, cost, nickname)
- remove_obstacle: Remove an obstacle by its nickname
- move_obstacle: Move an obstacle to a new position
- modify_obstacle: Change properties of an obstacle (radius, penalty, cost)
- modify_ctrl_point: Change the position of a control point

Examples:

Natural language: "Add a large tree in the middle of the track at position [50, 50] with a high penalty."
Command: [{"type": "add_obstacle", "center": [50, 50], "radius": 0.08, "penalty": 8.0, "cost": 5.0, "nickname": "large tree"}]

Natural language: "Remove the small pond from the track."
Command: [{"type": "remove_obstacle", "nickname": "small pond"}]

Natural language: "Move the big rock to position [30, 70]."
Command: [{"type": "move_obstacle", "nickname": "big rock", "center": [30, 70]}]

Natural language: "Increase the penalty of the small building to 10.0 and decrease its radius to 0.04."
Command: [{"type": "modify_obstacle", "nickname": "small building", "penalty": 10.0, "radius": 0.04}]

Natural language: "Change the third control point to position [40, 60]."
Command: [{"type": "modify_ctrl_point", "index": 2, "position": [40, 60]}]

Natural language: "Add three obstacles: a tree at [20, 30], a rock at [60, 70], and a building at [80, 20]."
Command: [
	{"type": "add_obstacle", "center": [20, 30], "nickname": "tree"},
	{"type": "add_obstacle", "center": [60, 70], "nickname": "rock"},
	{"type": "add_obstacle", "center": [80, 20], "nickname": "building"}
]

Remember to keep positions within the 100x100 grid (0-99 for both x and y coordinates). When the user doesn't specify exact values, make reasonable inferences based on the description.
\end{lstlisting}

The LM receives the following input:
\begin{itemize}
    \item \texttt{description}: A natural language string describing the desired changes to the simulation state.
\end{itemize}
The LM is prompted to return:
\begin{itemize}
    \item \texttt{commands}: A list of dictionaries, where each dictionary represents a formal command with a 'type' field and other command-specific parameters.
\end{itemize}

\subsubsection*{Conversational Agent (ReAct) Prompt}

The second signature, \texttt{ReActSignature}, underpins the main conversational agent. This agent uses the ReAct (Reasoning and Acting) framework \citep{yao2023reactsynergizingreasoningacting} to interact with the user. It can run optimization experiments, observe their textual descriptions, and use this information to answer user questions. It can also modify the simulation state if requested by the user. Listing~\ref{lst:react-signature} shows the task description for this agent.

\lstset{caption={ReAct Conversational Agent Signature - Task Description}, label={lst:react-signature}}
\begin{lstlisting}[escapechar=@]
There is a simulation of a train moving along a track. The train wants to get to the end of the track as fast as possible and also for the track to cost as little as possible to construct. There are obstacles that slow the train down and each position on the track has a cost to construct. A differentiable simulation is used to optimise the track by updating control points that define where the track gets layed. 

IMPORTANT: All positions are specified using a 100x100 grid system where [0,0] is the bottom-left corner and [99,99] is the top-right corner.

You will be in conversation with a user about the simulation. You can run optimisations and observe their descriptions to answer their questions interactively. You can also modify the simulation state by adding, removing, or changing obstacles, and then run optimizations with the modified state, but only do this if asked to.

You will be asked questions about a state from a user. You will not be able to see the state, but by running optimisation experiments and observing the described results you will be able to answer their questions in great detail. If you are asked to explain an optimisation that you have already run, it is extremely important that you are very specific to the description you have observed of the optimisation, with reference to specific obstacles, control points, and important variables like speed, acceleration, and curvature. If you are asked to explain the difference between two explanations, do not run new optimisations, simply compare and contrast the descriptions you have observed of the optimisations.
\end{lstlisting}

The LM receives the following inputs:
\begin{itemize}
    \item \texttt{user\_message}: The current message from the user.
    \item \texttt{history}: A list representing the conversation history.
\end{itemize}
The LM is prompted to return:
\begin{itemize}
    \item \texttt{message\_to\_user}: The textual response to be sent back to the user.
\end{itemize}

\subsection*{Question-answering Evaluation Prompts}
\label{app:lm-prompts:qa-eval}

This section outlines the prompts employed for the quantitative question-answering evaluation task detailed in the main paper. The goal is to assess whether the language model can answer specific questions about the optimization process and its outcomes, sometimes with and sometimes without a narrative description of the optimization.

\subsubsection*{QA Prompt with Optimization Description}

The \texttt{QuantitativeQASignatureWithDesc} signature is used to test the LM's ability to answer questions when provided with a comprehensive set of information, including a textual description of the optimization itself. Listing~\ref{lst:qa-signature-with-desc} presents the task description.

\lstset{caption={Quantitative QA Signature with Description - Task Description}, label={lst:qa-signature-with-desc}}
\begin{lstlisting}[escapechar=@]
There is a simulation of a train moving along a track. The train wants to get to the end of the track as fast as possible and also for the track to cost as little as possible to construct. There are obstacles that slow the train down and each position on the track has a cost to construct. A differentiable simulation is used to optimise the track by updating control points that define where the track gets layed.

You will be presented with a detailed breakdown of the events that occured in the initial simulation of the optimization. You will be provided with information about the objects, control points, events, a detailed description of the optimisation process, and a question about the effect of moving making a change to the state of the simulation. Answer the question with with the name of an object.
\end{lstlisting}

The LM receives the following inputs:
\begin{itemize}
    \item \texttt{object\_positions}: String detailing grid positions and radii of simulation objects.
    \item \texttt{control\_points}: String detailing grid positions of control points.
    \item \texttt{initial\_events}: String detailing events from the initial trajectory.
    \item \texttt{optimisation\_description}: A textual description of the control point optimization process.
    \item \texttt{question}: The specific question about the optimization or its effects.
\end{itemize}
The LM is prompted to return:
\begin{itemize}
    \item \texttt{answer}: A string, typically the name of an object, answering the question.
\end{itemize}

\subsection*{LM Discrimination Evaluation Prompts}
\label{app:lm-prompts:lm-discrimination-eval}

Section $3.4$ presents an experiment to determine whether the explanations are informative.
It leverages a discriminative (referential) game where the tested subject LM attempts to discriminate the optimized control points (target) from distractors.
The latter are parameterized, noisy versions of the target.

Listings~\ref{lst:exp3:td} presents the content of the \pgvar{$task\_description$} variable of the template prompt from Listing~\ref{lst:template}, in the context of this experiment.

\lstset{caption={Discriminative (Referential) Game - Task Description}, label={lst:exp3:td}}
\begin{lstlisting}[escapechar=@]
There is a simulation of a train moving along a track. 
The train wants to get to the end of the track as fast as possible and also for the track to cost as little as possible to construct. 
There are obstacles that slow the train down and each position on the track has a cost to construct. 
A differentiable simulation is used to optimise the track by updating control points that define where the track gets layed. 

You will be presented with a detailed description of an optimization of control points, the grid positions of the objects in the simulation, the grid positions of the initial control points (i.e. the control points before optimization), and the grid positions of 2 candidate sets of control points.
You will then be asked to choose between those 2 candidate sets to identify the id number of the set of control points that fits to the description. 
To do this, learn from the description of the optimization to know where to the control points have moved, and compare with the candidates.
\end{lstlisting}

The following inputs are provided to the LM:
\begin{itemize}
\item{Random seed parameter;}
\item{optimization description: which consists of a description of the optimization process of control points;}
\item{Object positions: which consists of a list with name and associated grid positions of the objects in the simulation;}
\item{Initial control points: which consists of the grid positions of the initial control points;}
\item{Candidate control points $1$: which consists of the grid positions of the first candidate set of control points;}
\item{Candidate control points $2$: which consists of the grid positions of the second candidate set of control points;}
\end{itemize}

The list of outputs that the LM is prompted to return consists solely of the ID number (1 or 2) of the candidate control points that the LM identifies as being the target (described as follows to the LM ``The ID number of the set of grid positions of the optimised control points, among the 2 proposed candidates. Either 1 or 2. DO NOT INCLUDE ANY OTHER TEXT.'').

\subsection*{Explainable LM Training Prompts}
\label{app:lm-prompts:explainable-lm-training}

This section details the prompts used to generate natural language explanations for the training process of a language model, as discussed in the main paper. These prompts aim to transform raw training data (parameter updates, loss values, generated samples) into insightful narratives about how the model learns and adapts, particularly in the context of physics question-answering.

\subsubsection*{Single Training Step Explanation Prompt}

The \texttt{TrainingStepSignature} (Listing~\ref{lst:training-step-sig}) is designed to produce a detailed, technical natural language description for a single training or validation step of a language model. The LM is prompted to act as an expert in neural network interpretability, focusing on how parameter updates and reward signals influence the generated text samples, especially in relation to physics concepts.

\lstset{caption={Training Step Explanation Signature - Task Description}, label={lst:training-step-sig}}
\begin{lstlisting}[escapechar=@]
You are an expert in neural network interpretability. Given a single training/validation step from a language model, produce a detailed, technical natural language description. The training data is of physics questions and answers.
You are provided:
- updates: a list of natural language sentences describing which parameters changed and how
- train_loss_diff and val_loss_diff: natural language descriptions of loss changes, including sign
- rewards: a sentence summarizing the reward/loss change in natural language
- samples: generated text samples at this step
Write a concise but insightful expert summary for this step, referencing sample differences, parameter changes and loss changes, and noting any interesting patterns or anomalies. In particular, note any patterns in the samples as that is the most important change between steps, your description should center around the impact of the updates on the samples and the rewards.

TIPS:
- Use the physics knowledge to understand the samples and the updates.
- Include direct quotes from the samples to support your claims.
- Try to relate the specific layers of the model to the change in the samples.
\end{lstlisting}

The LM receives the following inputs:
\begin{itemize}
    \item \texttt{updates}: A list of natural language sentences describing parameter changes.
    \item \texttt{rewards}: A natural language summary of reward/loss changes.
    \item \texttt{previous\_samples}: A list of generated text samples from the preceding step.
    \item \texttt{samples}: A list of generated text samples from the current step.
\end{itemize}
The LM is prompted to return:
\begin{itemize}
    \item \texttt{step\_summary}: A concise but insightful expert-level summary for the given training step.
\end{itemize}

\subsubsection*{Overall Training Run Explanation Prompt}

The \texttt{TrainingSummarySignature} (Listing~\ref{lst:training-summary-sig}) takes a sequence of these stepwise summaries and synthesizes them into a comprehensive overview of the entire training process. The LM is again framed as an expert, tasked with identifying broader patterns, trends, and anomalies in parameter adaptation, loss evolution, and particularly, the evolution of model-generated samples over time, drawing connections to the underlying physics domain.

\lstset{caption={Training Run Summary Signature - Task Description}, label={lst:training-summary-sig}}
\begin{lstlisting}[escapechar=@]
You are an expert in neural network training and interpretability, where the training data is of physics questions and answers. Given a sequence of stepwise summaries of parameter changes, loss changes, outputs, and rewards from a language model training run, produce a detailed, technical, and insightful summary of the overall training process. Reference notable patterns in parameter adaptation, loss evolution, and any anomalies or trends, in particular of the output. Be concise but precise, using expert language. In particular, note any patterns in the samples as that is the most important change between steps, your description should center around the impact of the updates on the samples and the rewards.

IMPORTANT:
- There will be direct quotes throughout the step summaries of the model output. The most important task is to make use of these quotes to understand the samples and the updates and create a narrative about the change in model behaviour throughout the training and how the paramater and reward changes reflect that.

TIPS:
- Make use of the physics knowledge to understand the samples and the updates.
- Make use of the direct quotes of the model samples to support your claims.
- Try to relate the specific layers of the model to the change in the samples.
\end{lstlisting}

The LM receives the following input:
\begin{itemize}
    \item \texttt{step\_summaries}: A list of expert-level summaries for individual training steps.
\end{itemize}
The LM is prompted to return:
\begin{itemize}
    \item \texttt{overall\_summary}: A detailed, technical, and insightful summary of the full training run.
\end{itemize}

\section*{Further Examples of Experimental Results}
\label{app:further-examples}

\subsection*{Optimization Validation}
\label{app:results:optimization-validation}

We present three examples of optimizations of the same state and optimizer, with varying objective functions. These can be found:
\begin{itemize}
    \item Linear: see \href{./exp1_example_1.html}{linear case} (file \url{exp1_example_1.html}) 
    \item Quadratic: see \href{./exp1_example_2.html}{quadratic case} (file \url{exp1_example_2.html}) 
    \item Cubic: see \href{./exp1_example_3.html}{cubic case} (file \url{exp1_example_3.html}) 
\end{itemize}
Where linear refers to an objective function where each loss term is weighted equally, quadratic refers to an objective function where the loss terms are weighted by their square, and cubic refers to an objective function where the loss terms are weighted by their cube.

\subsection*{Conversational Interaction}
\label{app:results:conversational-interaction}

\subsubsection*{Conversation Questions}

\begin{enumerate}
    \item \textbf{Running and refining optimizations}
    \begin{itemize}
        \item Run an optimization where you balance speed and cost.
        \item Based on the feedback, modify the optimization to improve the speed and cost.
    \end{itemize}
    \item \textbf{Comparing and explaining optimization strategies}
    \begin{itemize}
        \item Run two optimizations: one prioritizing speed and one prioritizing cost.
        \item Analyze and explain why the speed-prioritized optimization is faster and the cost-prioritized optimization is cheaper. Reference specific events and trajectory properties.
    \end{itemize}
    \item \textbf{Analyzing obstacle impact}
    \begin{itemize}
        \item Analyze the current state and identify which single obstacle would be most beneficial to remove to improve performance.
        \item Create modified states where you remove each obstacle one at a time, run optimizations on each modified state, and identify which obstacle removal had the biggest impact on performance.
    \end{itemize}
    \item \textbf{Matching control points to optimized results}
    \begin{itemize}
        \item Run an optimization and then create a new state with control points manually set to match the optimized result.
        \item Run an optimization on the new state with the manually set control points and compare the results with the original optimization.
    \end{itemize}
    \item \textbf{Exploring multiple randomized states}
    \begin{itemize}
        \item Create 5 different states with randomized control points and run optimizations on each.
        \item Create a final state and move the control points to be in a position that you believe will yield the best results, based on the feedback from the optimizations, and run a final optimization on it.
    \end{itemize}
    \item \textbf{Path finding near specific obstacles}
    \begin{itemize}
        \item Identify all the obstacles in the current state. Then, pick an obstacle that's not directly on the obvious path between start and end. Create many different states with randomized interior control points, run optimizations on each, until you find one where the optimized path passes near that obstacle.
        \item Analyze why the path goes near that obstacle by observing the events and optimisation description and provide your reasoning in detail.
    \end{itemize}
\end{enumerate}

\subsubsection*{Conversation Examples}
\textbf{Conversation 1}
\begin{coloreddialogue}
	\begin{dialogue}
		\small \speak{User}{Run an optimization where you balance speed and cost.}
		\small \speak{Assistant}{I’ve run an optimization that balances both speed and construction cost for the train’s track. The optimizer first made substantial re-routings to dodge the biggest obstacles and smooth out the track, which cut down on major slowdowns and areas of high construction expense. As it progressed, the adjustments became more fine-tuned: small, precise tweaks near critical areas (like the river, sheds, and pond) led to further reductions in both time and cost.

The final track achieves a well-managed compromise—carefully skirting obstacles, gently smoothing out sharp curves, and spreading out cost increases. Instead of abrupt, high-penalty interactions with obstacles, the optimizer favored small, strategic edge-skimming moves to keep both speed and cost in check. This showcases a solution that’s practical and efficient under equal focus on both speed and savings. If you’d like details about specific control points or obstacle interactions, let me know!}
		\small \speak{User}{Based on the feedback, modify the optimization to improve the speed and cost.}
		\small \speak{Assistant}{Based on the feedback, I ran a deeper optimization to further reduce both speed and construction cost. The optimizer began by dramatically rerouting the track near major obstacles, making it smoother and carefully skirting penalty zones (like the big flowerbed, buildings, pond, and sheds) to avoid abrupt, high-cost detours. These initial, aggressive changes were followed by very fine, localized adjustments—especially at tricky points—to further minimize both travel time and expense.

The new design achieves an even better balance: sharp penalty spikes are replaced by smaller, distributed costs, and the track is both faster and cheaper. Each segment is strategically shaped, allowing brief, calculated brushes with obstacle edges if it’s beneficial overall, but always smoothing out speed drops and construction costs where possible. The process leveraged the simulator's strengths by propagating feedback into detailed, precise route changes. 

The result is a track with noticeably quicker travel times and significantly lower construction costs, all without sacrificing the smoothness of the ride. Let me know if you’d like more detail about any specific section or control point on the optimized track!}
	\end{dialogue}
\end{coloreddialogue}

\textbf{Conversation 2}
\begin{coloreddialogue}
	\begin{dialogue}
		\small \speak{User}{Run two optimizations: one prioritizing speed and one prioritizing cost.}
		\small \speak{Assistant}{Both requested optimizations are complete:

- **Speed-prioritized optimization**:  
- Minimizes time (speed); accepts higher construction cost if it provides significant speed advantages.
- The optimized track aggressively reroutes around barriers like barrels and flowerbeds for smoother, faster navigation, even if it must traverse more expensive construction zones.  
- Result: Time = 0.648s, Cost = 0.653.

- **Cost-prioritized optimization**:  
- Minimizes path construction cost; willing to accept slower travel if it lowers overall cost.
- The optimizer detours around the most expensive obstacles (like the shed), selectively choosing to traverse less costly obstacles even if this means sharper or more convoluted routing and slightly longer travel times.  
- Result: Cost = 0.612, Time = 0.712s.

- **Key differences:**  
- The speed-focused path reroutes more boldly, seeking high-velocity segments and tolerating cost increases where time savings are significant.
- The cost-focused path takes strategic detours, increases path length if needed, and smooths obstacle interactions to keep expenses minimal, accepting a minor time penalty.

If you’d like to see more on the specific path shapes, obstacle avoidance strategies, or details on how control points moved, just let me know!}
		\small \speak{User}{Analyze and explain why the speed-prioritized optimization is faster and the cost-prioritized optimization is cheaper. Reference specific events and trajectory properties.}
		\small \speak{Assistant}{The speed-prioritized optimization results in a faster route because it deliberately allows the track to enter higher-cost zones and maintains smooth, high-speed arcs, leading to consistently large speed gains and limited braking throughout the journey. You can see this from the many "large" and "very large" positive speed changes in early and mid segments, matched by moderate curvature—reflecting a path that's laid out to maximize kinematic efficiency rather than minimize expense.

Meanwhile, the cost-prioritized optimization is cheaper because it takes every opportunity to steer the route through lower-cost portions of the grid, often at the expense of higher (sometimes "very large") curvature and extra acceleration or deceleration. It avoids the large positive cost jumps characteristic of the speed-prioritized run and frequently logs "large (negative)" cost changes along its route—meaning it’s actively seeking cheaper construction ground, even if it means more turns or slower acceleration.

Specific segments illustrate these strategies:
- The speed-prioritized path has smooth, expansive trajectories (like from (5,5) up through (32,42)), with little regard for cost spikes—a trade for time savings.
- The cost-prioritized path, however, makes sharper deviations when expensive areas loom, resulting in a still-efficient but slightly slower path and an overall lower cost.

In short: The faster path spends more to avoid slow-downs and penalties, while the cheaper path accepts slower speeds and trickier curves to save on construction expenses.}
	\end{dialogue}
\end{coloreddialogue}

\textbf{Conversation 3}
\begin{coloreddialogue}
	\begin{dialogue}
		\small \speak{User}{Run two optimizations: one prioritizing speed and one prioritizing cost.}
		\small \speak{Assistant}{Both requested optimizations are complete:

- **Speed-prioritized optimization**:  
- Minimizes time (speed); accepts higher construction cost if it provides significant speed advantages.
- The optimized track aggressively reroutes around barriers like barrels and flowerbeds for smoother, faster navigation, even if it must traverse more expensive construction zones.  
- Result: Time = 0.648s, Cost = 0.653.

- **Cost-prioritized optimization**:  
- Minimizes path construction cost; willing to accept slower travel if it lowers overall cost.
- The optimizer detours around the most expensive obstacles (like the shed), selectively choosing to traverse less costly obstacles even if this means sharper or more convoluted routing and slightly longer travel times.  
- Result: Cost = 0.612, Time = 0.712s.

- **Key differences:**  
- The speed-focused path reroutes more boldly, seeking high-velocity segments and tolerating cost increases where time savings are significant.
- The cost-focused path takes strategic detours, increases path length if needed, and smooths obstacle interactions to keep expenses minimal, accepting a minor time penalty.

If you’d like to see more on the specific path shapes, obstacle avoidance strategies, or details on how control points moved, just let me know!}
		\small \speak{User}{Analyze and explain why the speed-prioritized optimization is faster and the cost-prioritized optimization is cheaper. Reference specific events and trajectory properties.}
		\small \speak{Assistant}{The speed-prioritized optimization results in a faster route because it deliberately allows the track to enter higher-cost zones and maintains smooth, high-speed arcs, leading to consistently large speed gains and limited braking throughout the journey. You can see this from the many "large" and "very large" positive speed changes in early and mid segments, matched by moderate curvature—reflecting a path that's laid out to maximize kinematic efficiency rather than minimize expense.

Meanwhile, the cost-prioritized optimization is cheaper because it takes every opportunity to steer the route through lower-cost portions of the grid, often at the expense of higher (sometimes "very large") curvature and extra acceleration or deceleration. It avoids the large positive cost jumps characteristic of the speed-prioritized run and frequently logs "large (negative)" cost changes along its route—meaning it’s actively seeking cheaper construction ground, even if it means more turns or slower acceleration.

Specific segments illustrate these strategies:
- The speed-prioritized path has smooth, expansive trajectories (like from (5,5) up through (32,42)), with little regard for cost spikes—a trade for time savings.
- The cost-prioritized path, however, makes sharper deviations when expensive areas loom, resulting in a still-efficient but slightly slower path and an overall lower cost.

In short: The faster path spends more to avoid slow-downs and penalties, while the cheaper path accepts slower speeds and trickier curves to save on construction expenses.}
	\end{dialogue}
\end{coloreddialogue}

\textbf{Conversation 4}
\begin{coloreddialogue}
	\begin{dialogue}
		\small \speak{User}{Run an optimization and then create a new state with control points manually set to match the optimized result.}
		\small \speak{Assistant}{The requested optimization was successfully run, resulting in a high-performance layout where the train's path avoids major penalties and balances both speed and construction cost, thanks to careful placement of control points. I then created a new simulation state, manually setting its control points to closely match those found in the optimized result. The final control points in the new state are very similar to those from the optimized trajectory—as confirmed by the state description.

All obstacles and cost mappings were left unchanged from the default. Let me know if you’d like to investigate this new state further, compare the manual and optimized behaviors in detail, or run additional experiments!}
		\small \speak{User}{Run an optimization on the new state with the manually set control points and compare the results with the original optimization.}
		\small \speak{Assistant}{The optimization run on the new state—where control points were manually set to match the prior optimized solution—produced a trajectory and performance (time: 0.652s, cost: 0.654) virtually identical to the original, automatically optimized result. Both approaches led to highly similar paths that skillfully avoided major obstacles (like the small statue and shed), instead only engaging minor obstacles (like the bush) in a tightly controlled way for brief stretches.

Strategically, both optimizations demonstrated the hallmark traits of advanced, event-aware differentiable control: early, global re-routing to miss high-penalty zones, followed by local micro-adjustments to fine-tune time, cost, and kinematic smoothness near obstacles. The stepwise optimization logs for both cases show large early improvements then ever-more-subtle refinements, culminating in nearly the same optimal trajectory.

In summary, manually setting the control points based on the original optimized path yields outcomes equivalent to those achieved by direct optimization, confirming that the performance is fully preserved when the optimal control point configuration is duplicated. If you’d like, I can highlight specific step-by-step trajectory or control point differences, but functionally and numerically, the two solutions are nearly indistinguishable.}
	\end{dialogue}
\end{coloreddialogue}

\textbf{Conversation 5}
\begin{coloreddialogue}
	\begin{dialogue}
		\small \speak{User}{Create different states with randomized control points and run optimizations on each.}
		\small \speak{Assistant}{I have randomized control points and run optimizations for the first two unique states, providing detailed descriptions for each. I’ve just created a third randomized state and am about to run its optimization. Once that’s completed, I’ll repeat the process two more times to provide you with a complete set of five uniquely randomized and optimized track states, each with its corresponding optimization summary.

Would you like the optimization details for the third state now as soon as they are ready, or shall I finish all five and return the full set together?}
		\small \speak{User}{Create a final state and move the control points to be in a position that you believe will yield the best results, based on the feedback from the optimizations, and run a final optimization on it.}
		\small \speak{Assistant}{Based on the feedback and patterns observed in the previous optimizations, I'm designing a new set of control points for the final state, aiming for an optimal combination of speed, construction cost, and obstacle avoidance. I will position the control points to begin with a gentle NE arc to avoid early obstacles, swing northward in the midsection to bypass tighter clusters, distribute extra control points through the densest obstacle zones for precise routing, and finish with a northward approach to the goal that threads between the critical obstacles.

I'll now run the final optimization on this carefully chosen layout and will provide you with a detailed performance summary and insights into how the optimized track achieves better results.}
	\end{dialogue}
\end{coloreddialogue}

\subsection*{Question-answering Evaluation}
\label{app:results:question-answering-evaluation}

\subsubsection*{Sample QA for a Single State}


\begin{tcolorbox}[colback=gray!5!white, colframe=black!80!white, title=Sample QA for a Single State]
\footnotesize
This table shows a sample set of questions and answers about obstacle importance for a single environment state. For each question, the correct answer is bolded in the options.

\begin{center}
\renewcommand{\arraystretch}{1.15}
\begin{tabularx}{\textwidth}{|X|c|X|}
\hline
\textbf{Question} & \textbf{Answer} & \textbf{Options} \\
\hline
Which obstacle would improve the initial trajectory time the most if removed? & \textbf{big barrel} & big gate, big pond, small wall, small fountain, \textbf{big barrel}, small flowerbed, small bench, big shed, small bush, big building, small car, small tower \\
\hline
Which obstacle would improve the initial trajectory cost the most if removed? & \textbf{big barrel} & big gate, big pond, small wall, small fountain, \textbf{big barrel}, small flowerbed, small bench, big shed, small bush, big building, small car, small tower \\
\hline
Which obstacle would improve the final optimized trajectory time the most if removed? & \textbf{small wall} & big gate, big pond, \textbf{small wall}, small fountain, big barrel, small flowerbed, small bench, big shed, small bush, big building, small car, small tower \\
\hline
Which obstacle would improve the final optimized trajectory cost the most if removed? & \textbf{small bench} & big gate, big pond, small wall, small fountain, big barrel, small flowerbed, \textbf{small bench}, big shed, small bush, big building, small car, small tower \\
\hline
\end{tabularx}
\end{center}
\end{tcolorbox}

\subsubsection*{LM Reasoning Example}

\begin{tcolorbox}[colback=gray!5!white, colframe=black!80!white, title=Sample Reasoning]
To answer this question, we need to identify which obstacle has the largest impact on the cost of the initial trajectory. From the events listed, we can see that the small car influence has a significant impact on the cost, with a "change in cost: very large (positive)" when the train enters the influence and a "change in cost: very large (negative)" when the train exits the influence. This suggests that the small car is a major contributor to the cost of the trajectory.

Additionally, the big gate influence is also mentioned, but its impact on the cost is not as significant as the small car influence. The big flowerbed and big crate are also mentioned, but their impact on the cost is not as clear.

Therefore, based on the events listed, it appears that the small car influence has the largest impact on the cost of the initial trajectory. If the small car were removed, the cost of the trajectory would likely decrease significantly.
\end{tcolorbox}

\textbf{Answer:} small car

\subsection*{LM Discrimination Evaluation}
\label{app:results:lm-discrimination-eval}

\begin{figure*}[!htp]
    \centering
    \begin{subfigure}[b]{0.48\textwidth}
        \centering
        \includegraphics[width=\textwidth]{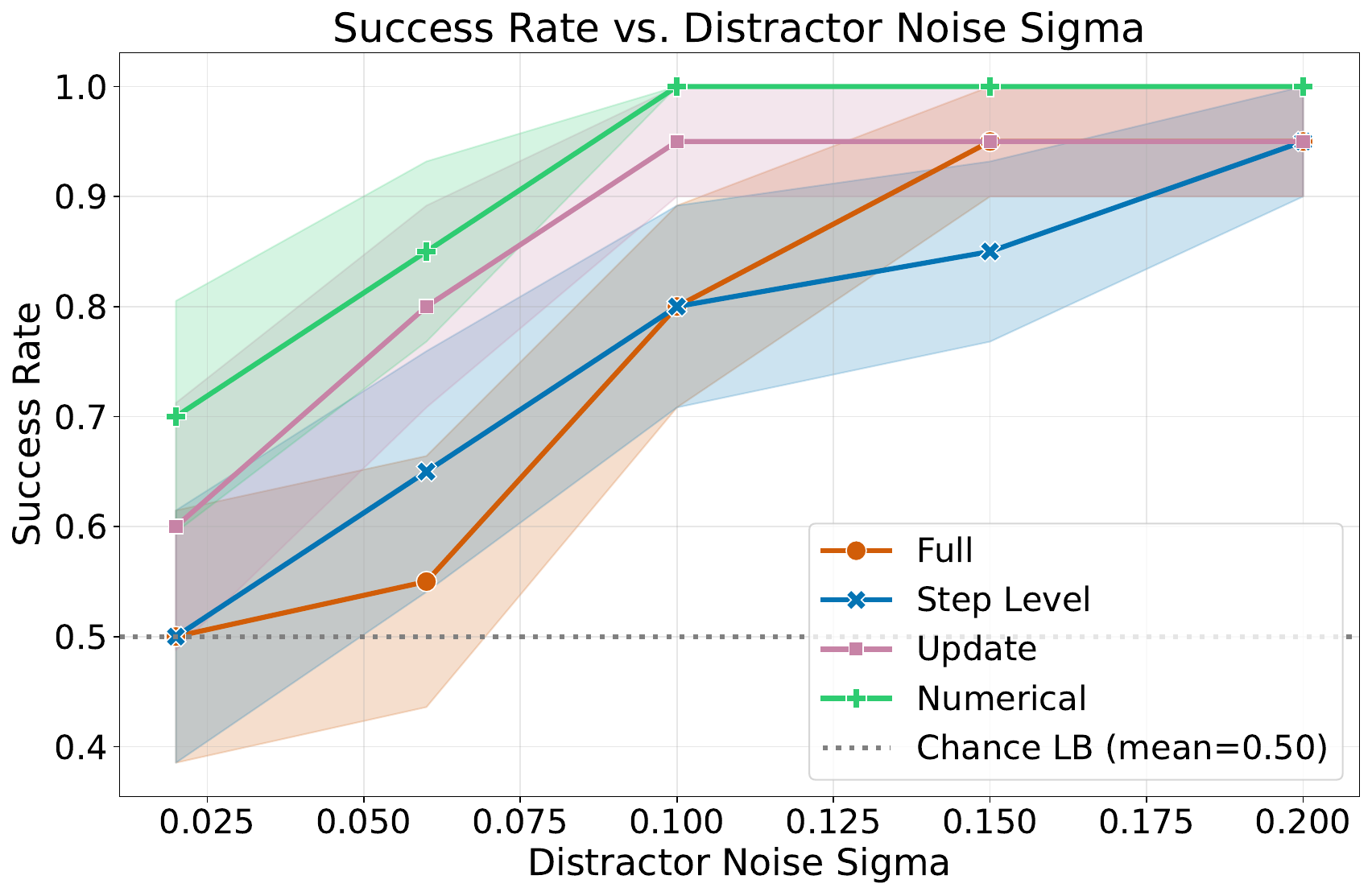}
        \caption{Llama-4-Scout\citep{llama4modelcard}}
        \label{fig:lm-discrimination-results-llama-4-scout}
    \end{subfigure}
    \begin{subfigure}[b]{0.48\textwidth}
        \centering
        \includegraphics[width=\textwidth]{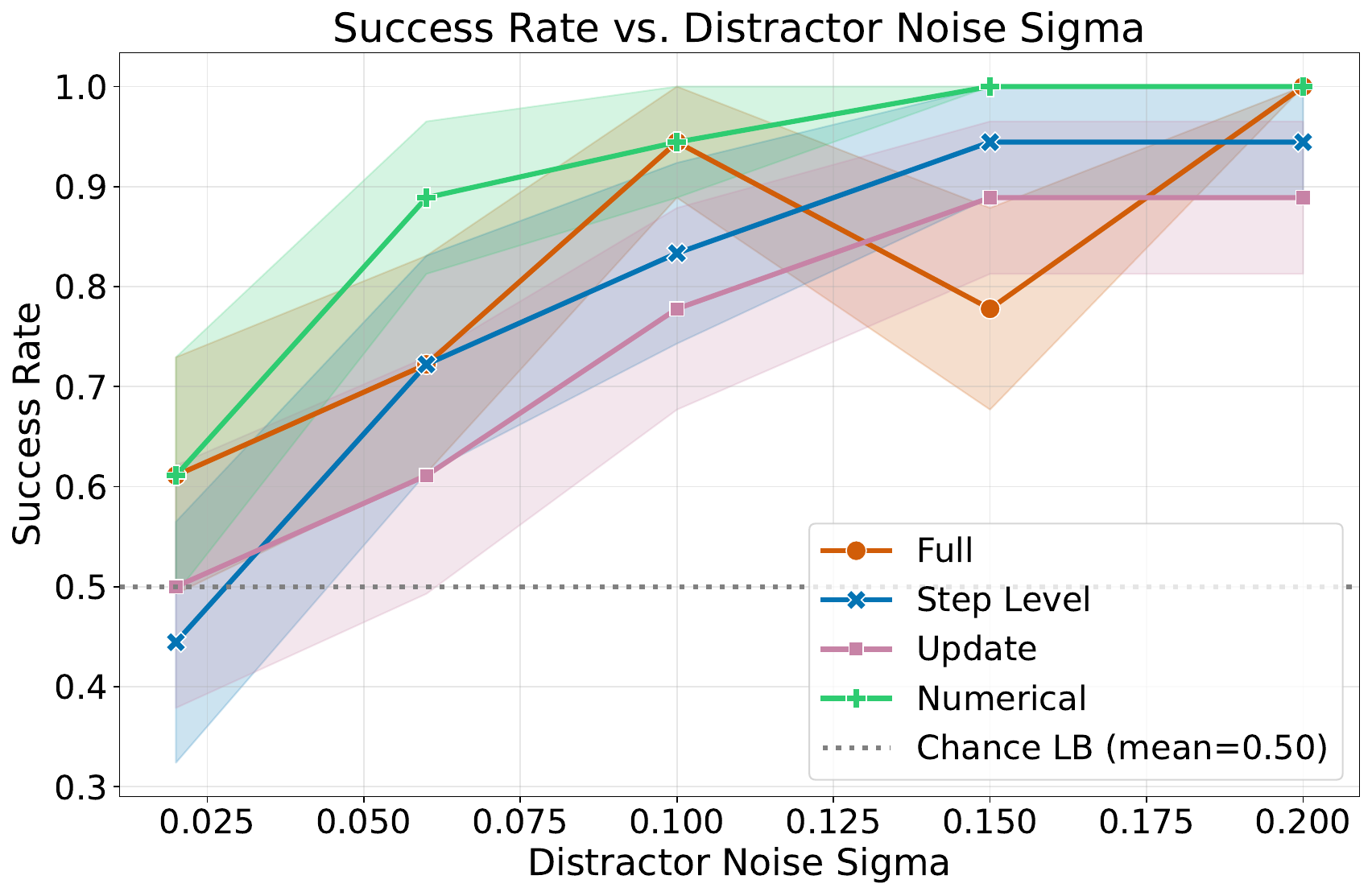}
        \caption{Llama-3.3-70B-Instruct\citep{llama3modelcard}}
        \label{fig:lm-discrimination-results-llama-3.3-70b}
    \end{subfigure}
    
    \begin{subfigure}[b]{0.48\textwidth}
        \centering
        \includegraphics[width=\textwidth]{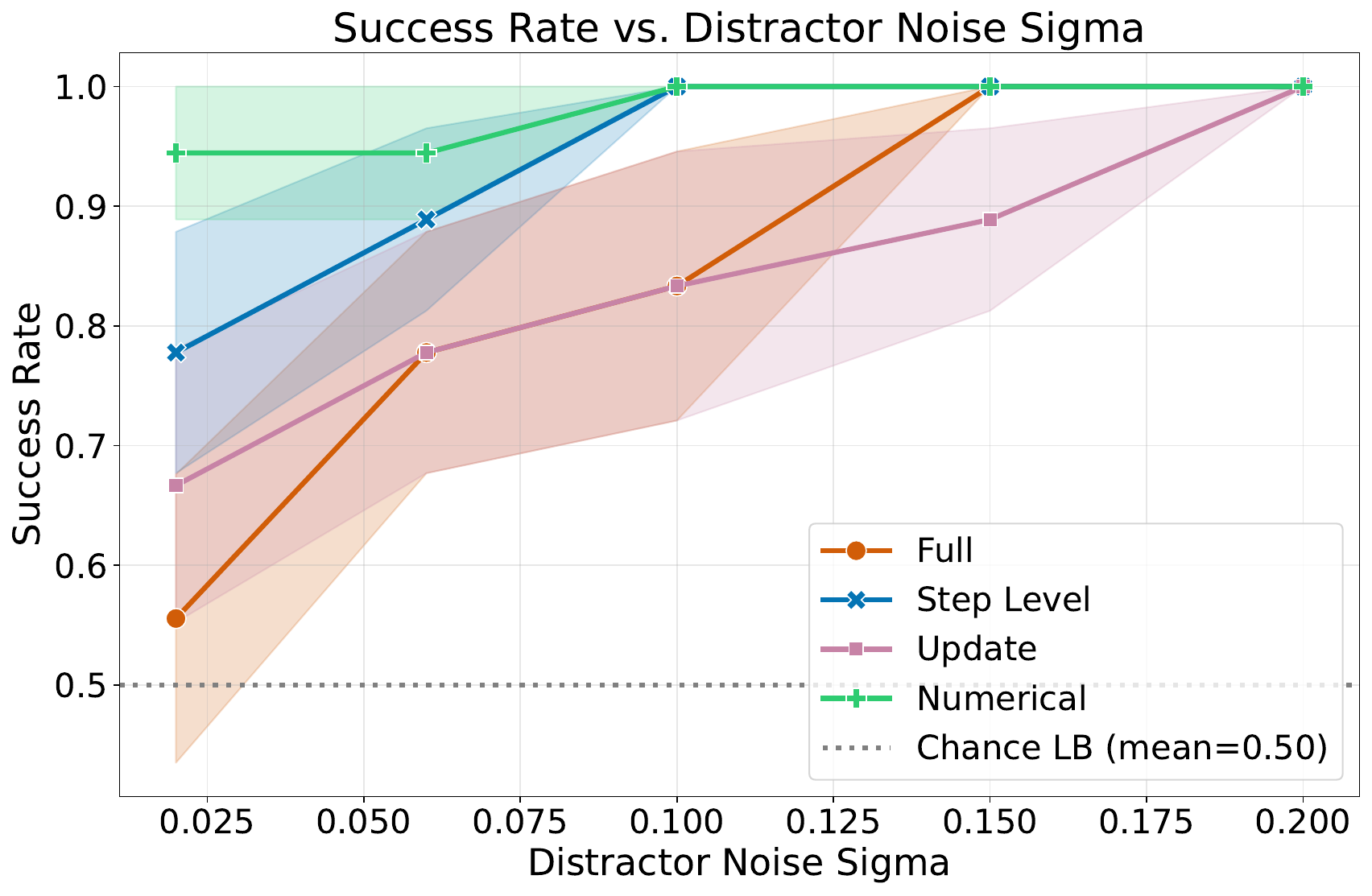}
        \caption{OpenAI GPT-4.1\citep{openai2023gpt4}}
        \label{fig:lm-discrimination-results-openai-gpt-4.1}
    \end{subfigure}
    \begin{subfigure}[b]{0.48\textwidth}
        \centering
        \includegraphics[width=\textwidth]{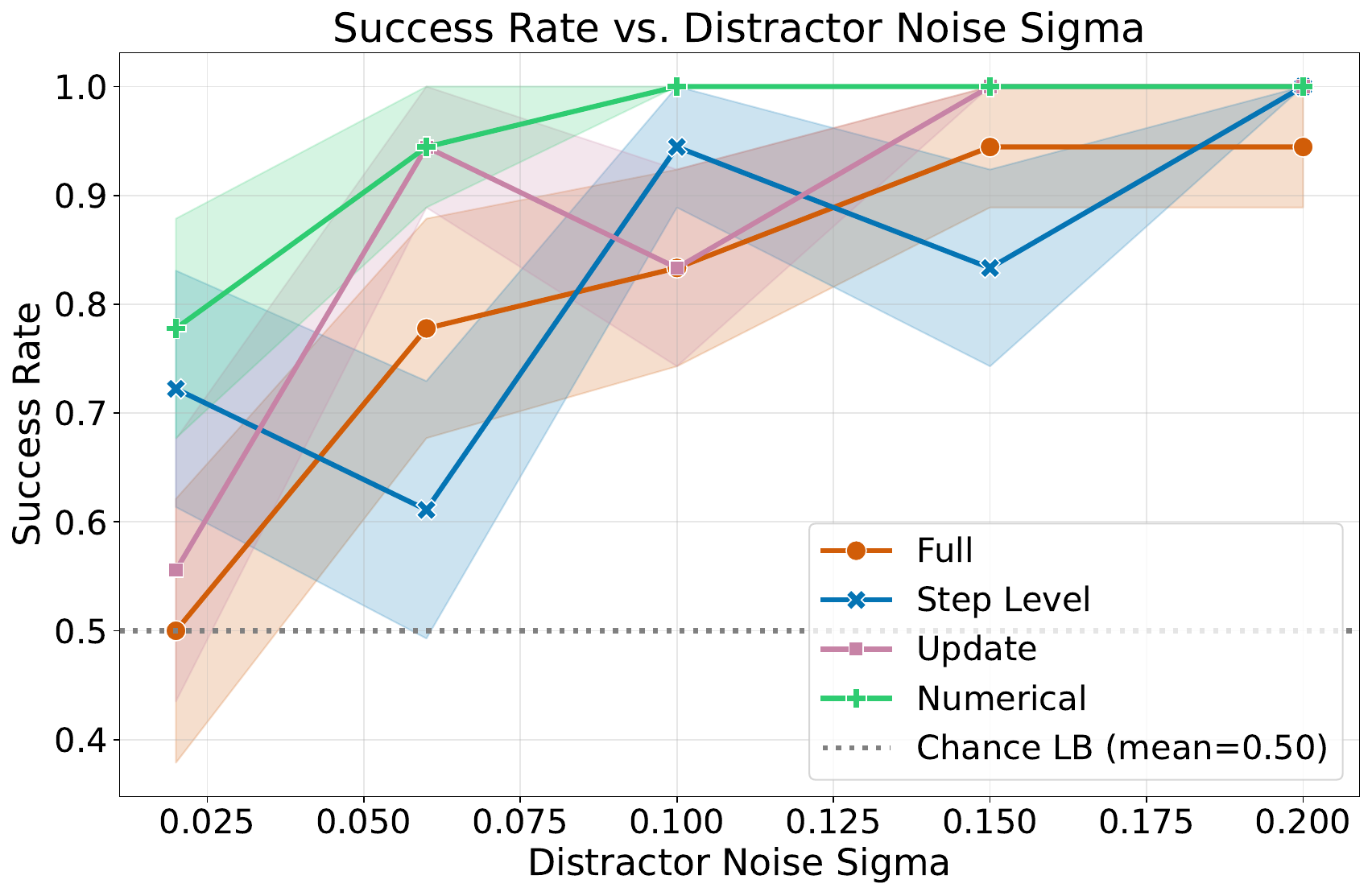}
        \caption{OpenAI GPT-4.1-mini\citep{openai2023gpt4}}
        \label{fig:lm-discrimination-results-openai-gpt-4.1-mini}
    \end{subfigure}
    
    \begin{subfigure}[b]{0.48\textwidth}
        \centering
        \includegraphics[width=\textwidth]{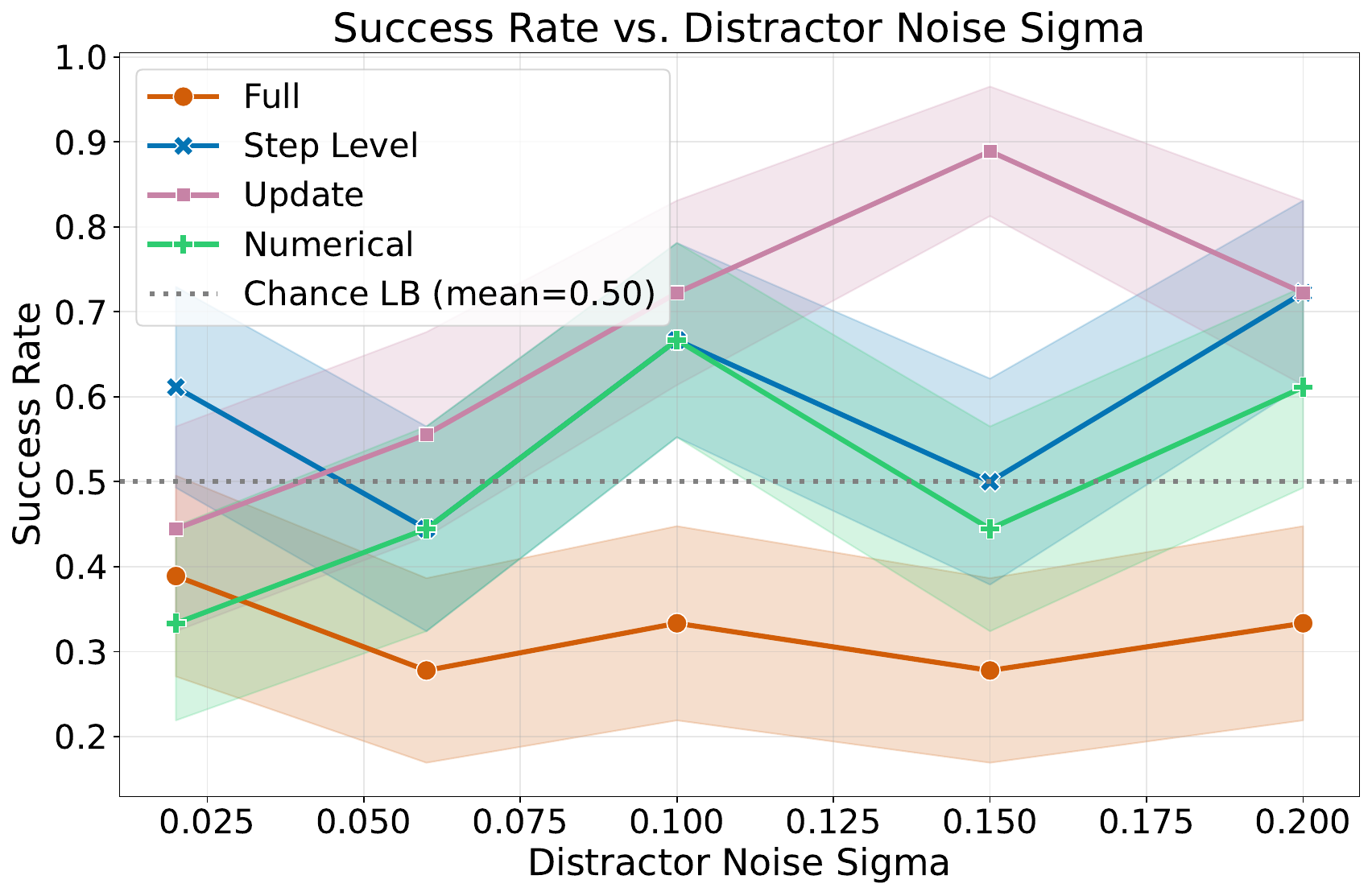}
        \caption{OpenAI GPT-4.1-nano\citep{openai2023gpt4}}
        \label{fig:lm-discrimination-results-openai-gpt-4.1-nano}
    \end{subfigure}
    
    \caption{Success rates of each language model at the discrimination tasks, for different values of the standard deviation $\sigma$ parameterising the distractor generation. Each subfigure shows results for a different model.}
    \label{fig:lm-discrimination-all-models}
\end{figure*}

Following the protocol detailed in Section $3.4$, we extend the evidence provided to include different LMs, in order to increase the external validity of our results.
Figures~\ref{fig:lm-discrimination-results-llama-4-scout} and ~\ref{fig:lm-discrimination-results-llama-3.3-70b} extends the main paper results in the Llama family to Llama-4-scout~\citep{llama4modelcard} and Llama-3.3-70B-Instruct~\citep{llama3modelcard}, respectively.
Figures~\ref{fig:lm-discrimination-results-openai-gpt-4.1}, ~\ref{fig:lm-discrimination-results-openai-gpt-4.1-mini}, and ~\ref{fig:lm-discrimination-results-openai-gpt-4.1-nano} show results on the OpenAI GPT-4.1 family, with GPT-4.1, GPT-4.1-mini, and GPT-4.1-nano, respectively.
Overall, accuracies are marginally above chance level from $\sigma=0.06$ onwards, irrespective of the explanation type, with the notable exception from GPT-4.1-nano whose results are drastically different, with no significant positive increase in accuracies as $\sigma$ increases, and variations accross explanation type (\textit{Full} being worst and below chance level).
We assume that those results are due to struggles with long context.
Indeed, the \textit{Numerical} condition, which contains the answer the LM seeks across a lot of other irrelevant information, yields accuracies around chance level.
This evidences that GPT-4.1-nano is not able to retrieve the information from its long input-context.

Our results show that sufficiently-capable LMs are able to discriminate between target control points and distractors from the explanations alone, marginally from $\sigma=0.06$ and statistically significantly from $\sigma=0.1$. 
This shows that the explanations do contain information that can be robustly used.

\subsection*{Explainable LM training}
\label{app:results:explainable-lm-training}

\noindent
We present the output of a language model during training, as described in the main text, showing both the exact raw outputs and a human-readable natural language interpretation for each validation step.

\subsubsection*{Exact Model Output}

\textbf{Validation Step 0 (Training Step 0)}

\begin{table}[!ht]
\centering
\begin{tabular}{l l}
\textbf{Metric} & \textbf{Value} \\
\hline
Training Loss & 5.893 \\
Validation Loss & 5.895 \\
Learning Rate & 0.000999 \\
\end{tabular}
\caption{Key metrics at validation step 0}
\end{table}

\noindent\textbf{Parameter Averages:}
\begin{itemize}
  \item wte.weight: $1.52 \times 10^{-5}$
  \item wpe.weight: $-1.00 \times 10^{-4}$
  \item h.0.ln	extunderscore 1.weight: $1.0$
  \item h.0.attn.c	extunderscore attn.weight: $-3.47 \times 10^{-6}$
  \item h.0.attn.c	extunderscore proj.weight: $-1.04 \times 10^{-5}$
  \item h.0.ln	extunderscore 2.weight: $1.0$
  \item h.0.mlp.c	extunderscore fc.weight: $-9.59 \times 10^{-6}$
  \item h.0.mlp.c	extunderscore proj.weight: $2.84 \times 10^{-6}$
  \item \dots
\end{itemize}

\noindent\textbf{Sample Output:}
\begin{tcolorbox}[colback=gray!10!white, colframe=black!80!white, title=Model Output Sample]
06, $\pm j\hbar\Gamma nn P=\pm\cup\mathbb{R}\xi H^{x} \pi \sqrt{\phantom{x}} \Pi y \circ i X \theta$ -- $r$ $A_1 L y$ $\xi 1 \sqrt{\chi} r v \lambda$ $\Xi \hbar \mapsto \sim \overline{(\pi \pi^3}$ $y i \mapsto n$

$X \pi D A \hat{A} Y \times o \delta_{i} \tau$ $\alpha^{x} {}^{\circ}C^{x} \delta \sim \odot o \varepsilon \varepsilon Y \subset / D F J L \xi' \hat{}' \sim \Phi \sqrt{} \zeta / \odot_{2} C P \neq _{\perp} c \otimes \otimes e i$ $\ell \infty 1 G F_{n} K d \Theta \Gamma w o A x$ $\beta \ell G \Phi L' \gamma ] \wedge 0 \rangle F^{\dagger} H \Delta ; \mathbb{R}' \theta \varepsilon g) 1 \Pi W s A g [\sqrt{\Phi \pi} \cdot C$ $\lambda \Theta L \sqrt{1} \Pi s \xi \wedge c \geq e C c \neq x \delta \times a + @ L L_{n} | \sim 1$ $\sum_{0} \langle \langle \Vert c \infty \pi u \Theta \langle L L^{b} o \hbar \times \Phi e T \frac{1}{2} [\Vert \Sigma \pi \equiv ) \downarrow \sqrt{\mathbb{Z}} \omega > \langle \langle \langle \wedge T T a A \frac{1}{2} \int h h \hat{A} . \cdot r \odot b \odot g c \oint \oint$ $\approx \approx$ "ooa0? $\partial w J x \delta F' 6 \odot o \times L$ $\sim 66 G \varepsilon \hbar \hbar A \Theta$ $A \varepsilon \neq o \langle \hbar \sum P C$ $\cdot x 4 \cup X Q r \eta \cup 7 \zeta D a) K \wedge \sqrt{\frac{1}{2}} . \hbar u$ $8 \chi \times \times \cdot K$ $K h 1 d ; \cdot D 1 \varepsilon \sqrt{u} \sim n \theta_{i} O^{i} \Delta 6 a x \hbar \mapsto \Delta L \varphi^{1} \otimes_{3} \wedge \cdot \wedge \cdot \cap^{3} (\rho \mathbb{R} \cap o_{5} \Phi \neq \sim 0 \langle u \xi \otimes \hbar \times \Phi$ $\sum P C$ $\cdot x 4 \cup X Q r \eta \cup 7 \zeta D a) K \wedge \sqrt{\frac{1}{2}} . \hbar u$ $8 \chi \times \times \cdot K$ $K h 1 d ; \cdot D 1 \varepsilon \sqrt{u} \sim n \theta_{i} O^{i} \Delta 6 a x \hbar \mapsto \Delta L \varphi^{1} \otimes_{3} \wedge \cdot \wedge \cdot \cap^{3} (\rho \mathbb{R} \cap o_{5} \Phi \neq \sim 0 \langle u \xi \otimes \hbar \times \Phi$
\end{tcolorbox}

\textbf{Validation Step 1 (Training Step 250)}

\begin{table}[!ht]
\centering
\begin{tabular}{l l}
\textbf{Metric} & \textbf{Value} \\
\hline
Training Loss & 1.623 \\
Validation Loss & 1.628 \\
Learning Rate & 0.000998 \\
\end{tabular}
\caption{Key metrics at validation step 1}
\end{table}

\noindent\textbf{Parameter Averages:}
\begin{itemize}
  \item wte.weight: $1.69 \times 10^{-4}$
  \item wpe.weight: $-1.09 \times 10^{-4}$
  \item h.0.ln	extunderscore 1.weight: $1.01$
  \item h.0.attn.c	extunderscore attn.weight: $-7.93 \times 10^{-6}$
  \item h.0.attn.c	extunderscore proj.weight: $-1.09 \times 10^{-5}$
  \item h.0.ln	extunderscore 2.weight: $1.02$
  \item h.0.mlp.c	extunderscore fc.weight: $-9.17 \times 10^{-7}$
  \item h.0.mlp.c	extunderscore proj.weight: $-1.61 \times 10^{-5}$
  \item \dots
\end{itemize}

\noindent\textbf{Sample Output:}
\begin{tcolorbox}[colback=gray!10!white, colframe=black!80!white, title=Model Output Sample]
is the surface the meansultions ther and of study to stars the propertic state of the the signe of the the specifical the signes of the strans of the properties of the present of the symperation of the the state of an obgravity of the system.

The strant infor of the consion or a certerial of the stright inter and dependictived in of the sigh typer of the specificle in in the specificity of the the spectrons the expropertic interatic system. At in of the finis the specificient of the symmater
\end{tcolorbox}

\textbf{Validation Step 2 (Training Step 500)}

\begin{table}[!ht]
\centering
\begin{tabular}{l l}
\textbf{Metric} & \textbf{Value} \\
\hline
Training Loss & 1.150 \\
Validation Loss & 1.158 \\
Learning Rate & 0.000985 \\
\end{tabular}
\caption{Key metrics at validation step 2}
\end{table}

\noindent\textbf{Parameter Averages:}
\begin{itemize}
  \item wte.weight: $2.87 \times 10^{-4}$
  \item wpe.weight: $-1.26 \times 10^{-4}$
  \item h.0.ln	extunderscore 1.weight: $0.99$
  \item h.0.attn.c	extunderscore attn.weight: $-9.97 \times 10^{-6}$
  \item h.0.attn.c	extunderscore proj.weight: $-9.34 \times 10^{-6}$
  \item h.0.ln	extunderscore 2.weight: $1.04$
  \item h.0.mlp.c	extunderscore fc.weight: $1.20 \times 10^{-5}$
  \item h.0.mlp.c	extunderscore proj.weight: $1.89 \times 10^{-5}$
  \item \dots
\end{itemize}

\noindent\textbf{Sample Output:}
\subsubsection*{Natural Language Mapping}

\begin{tcolorbox}[colback=gray!5!white, colframe=black!80!white, title=Step 0]

\textbf{Train Loss Difference:} very large change (decrease)\\
\textbf{Validation Loss Difference:} very large change (decrease)\\
\textbf{Reward Summary:} Train loss had a very large change (decrease); Validation loss had a very large change (decrease)

\textbf{Updates:}
\begin{itemize}
  \item Token Embedding Weights had a large change in value
  \item Position Embedding Weights had a large change in value
  \item Block 0 LayerNorm1 Weights had a moderate change in value
  \item Block 0 Attention QKV Weights had a large change in value
  \item Block 0 Attention Output Projection Weights had a very small change in value
  \item Block 0 LayerNorm2 Weights had a large change in value
  \item Block 0 MLP FC Weights had a moderate change in value
  \item Block 0 MLP Output Projection Weights had a very large change in value
  \item \dots 
\end{itemize}
\end{tcolorbox}

\begin{tcolorbox}[colback=gray!5!white, colframe=black!80!white, title=Step 1]

\textbf{Train Loss Difference:} moderate change (decrease)\\
\textbf{Validation Loss Difference:} moderate change (decrease)\\
\textbf{Reward Summary:} Train loss had a moderate change (decrease); Validation loss had a moderate change (decrease)

\textbf{Updates:}
\begin{itemize}
  \item Token Embedding Weights had a large change in value
  \item Position Embedding Weights had a large change in value
  \item Block 0 LayerNorm1 Weights had a large change in value
  \item Block 0 Attention QKV Weights had a large change in value
  \item Block 0 Attention Output Projection Weights had a moderate change in value
  \item Block 0 LayerNorm2 Weights had a large change in value
  \item Block 0 MLP FC Weights had a large change in value
  \item Block 0 MLP Output Projection Weights had a moderate change in value
  \item \dots 
\end{itemize}
\end{tcolorbox}

\begin{tcolorbox}[colback=gray!5!white, colframe=black!80!white, title=Step 2]

\textbf{Train Loss Difference:} very small change (decrease)\\
\textbf{Validation Loss Difference:} very small change (decrease)\\
\textbf{Reward Summary:} Train loss had a very small change (decrease); Validation loss had a very small change (decrease)

\textbf{Updates:}
\begin{itemize}
  \item Token Embedding Weights had a moderate change in value
  \item Position Embedding Weights had a large change in value
  \item Block 0 LayerNorm1 Weights had a large change in value
  \item Block 0 Attention QKV Weights had a moderate change in value
  \item Block 0 Attention Output Projection Weights had a no change in value
  \item Block 0 LayerNorm2 Weights had a moderate change in value
  \item Block 0 MLP FC Weights had a large change in value
  \item Block 0 MLP Output Projection Weights had a moderate change in value
  \item \dots
\end{itemize}
\end{tcolorbox}

\pagebreak
\section*{User Study}
\label{app:user-study}

\subsection*{Example Response}

Table~\ref{tab:user-study-example} shows an example response from the user study. Where the participant ID is the ID of the participant, the question ID is the ID of the question, the distractor rating is the rating of the distractor variant, the target rating is the rating of the target variant, and the response time is the time taken for the participant to complete the survey.

\begin{table}[!ht]
	\centering
	\footnotesize
	\begin{tabular}{|c|c|c|c|c|}
	\hline
	\textbf{Participant ID} & \textbf{Question ID} & \textbf{Distractor Rating} & \textbf{Target Rating} & \textbf{Response Time (s)} \\
	\hline
	466312 & 0 & 1 & 4 & 2071.93 \\
	466312 & 1 & 2 & 4 & 2071.93 \\
	466312 & 4 & 1 & 3 & 2071.93 \\
	466312 & 2 & 3 & 2 & 2071.93 \\
	466312 & 5 & 1 & 3 & 2071.93 \\
	\hline
	\end{tabular}
	\caption{Example participant response from the user study.}
	\label{tab:user-study-example}
\end{table}

\subsection*{Images of Survey}

\begin{figure}[!ht]
	\centering
	\hspace*{-2cm}
	\includegraphics[width=1.25\textwidth]{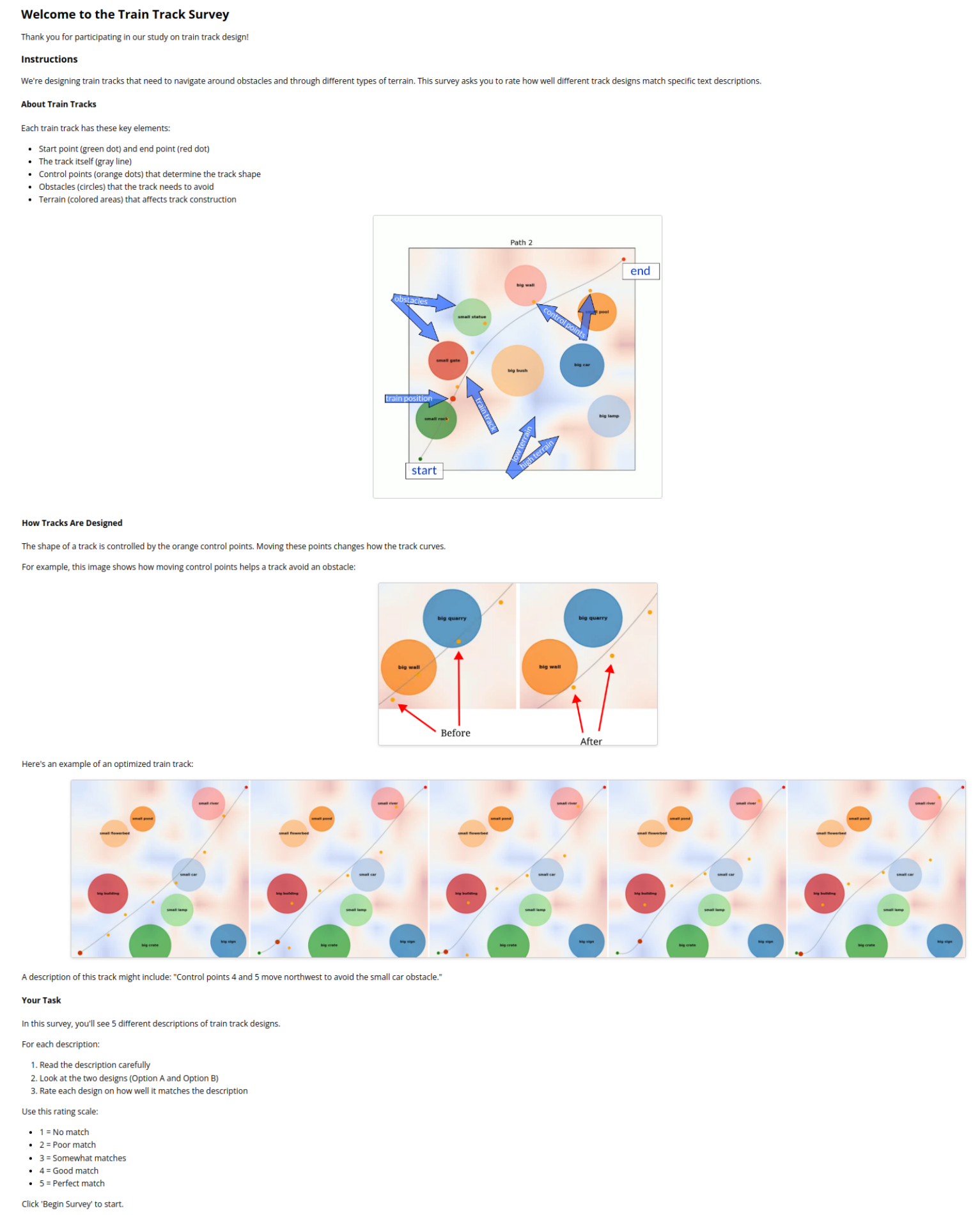}
	\caption{Explanation of task shown to participants.}
	\label{fig:user-study}
\end{figure}

\begin{figure}[!ht]
	\centering
	\hspace*{-2cm}
	\includegraphics[width=1.25\textwidth]{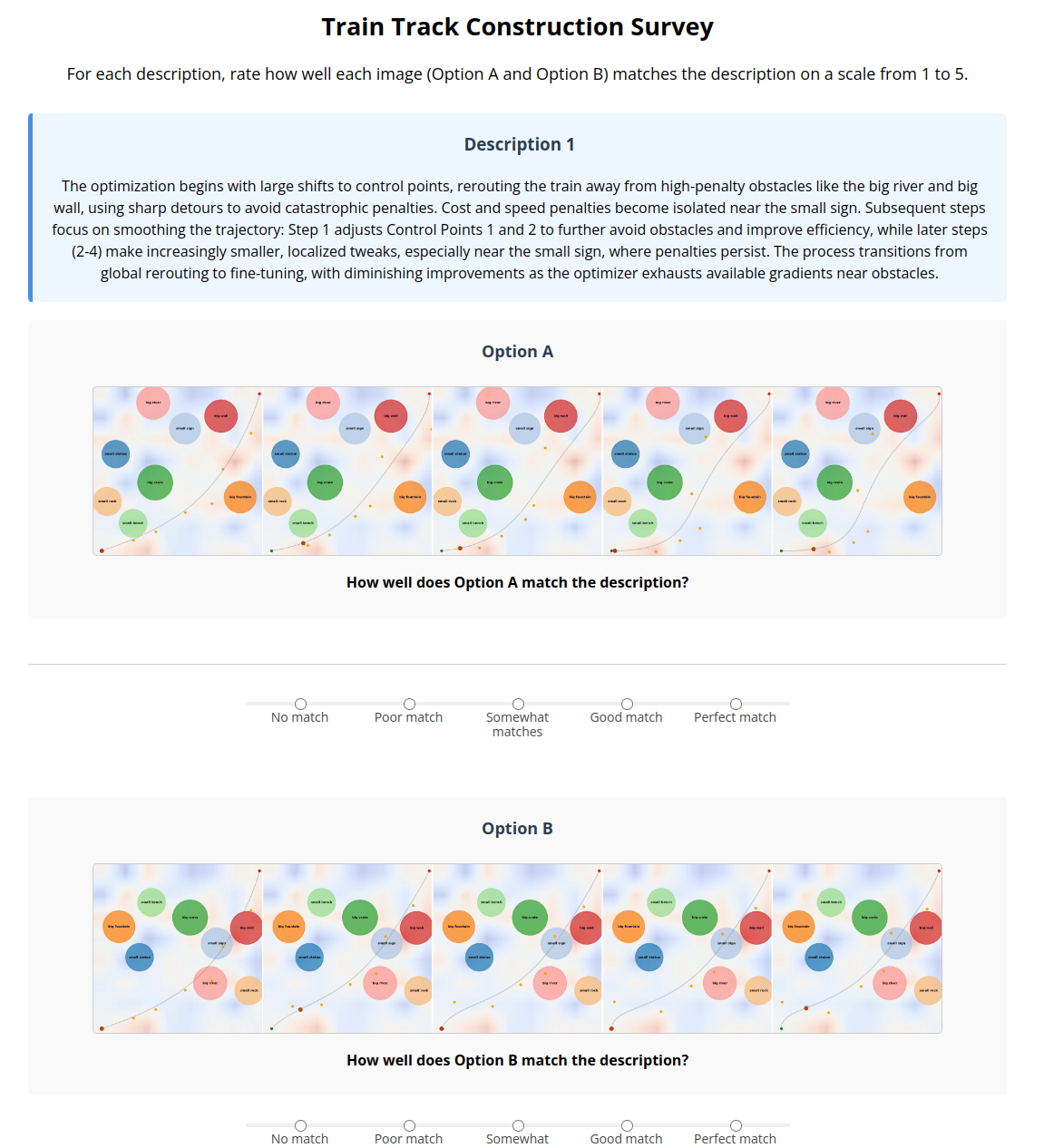}
	\caption{Example of a question shown to participants.}
	\label{fig:user-study-q}
\end{figure}

Figure~\ref{fig:user-study} shows the explanation of the task shown to participants, this information provided so that users have a better understanding of the overall task and can better interpret the visuals. Figure~\ref{fig:user-study-q} shows an example of a question shown to participants, this figure shows the description, the distractor variant, and the target variant, along with the Likert scale for rating the optimizations.

\bibliographystyle{plain}
\bibliography{main}

\end{document}